\documentclass[10pt,twocolumn,letterpaper]{article}

\usepackage{cvpr}
\usepackage{times}
\usepackage{epsfig}
\usepackage{graphicx}
\usepackage{amsmath}
\usepackage{amssymb}
\usepackage{url}
\usepackage{epsfig}
\usepackage{graphicx}
\usepackage{amsmath}
\usepackage{amssymb}
\usepackage{amsfonts}    
\usepackage{nicefrac}       % compact symbols for 1/2, etc.
\usepackage{microtype}      % microtypography
\usepackage{xcolor}
\usepackage{grffile}
\usepackage{amsmath}
\usepackage{graphicx}
\usepackage{verbatim}
\usepackage{setspace}
\usepackage{subcaption}
\usepackage{verbatim}
\usepackage{multirow}
\usepackage{mwe}
\usepackage{verbatimbox}
\usepackage{array,booktabs}
\usepackage[inline]{enumitem}
\usepackage{mathtools}
\usepackage{floatrow}
\usepackage{multirow}
\usepackage{placeins}
\usepackage{array}
\usepackage[pagebackref=true,colorlinks=false]{hyperref}
\usepackage{hypcap}
\newcolumntype{C}{>{$\displaystyle}c<{$}}

\usepackage[title,toc,titletoc,page]{appendix}

\newcommand{\myparagraph}[1]{\vspace{0.0em}\noindent\textbf{#1}}
\newfloatcommand{capbtabbox}{table}[][\FBwidth]
\newcolumntype{C}[1]{>{\centering\let\newline\\\arraybackslash\hspace{0pt}}m{#1}}

\DeclarePairedDelimiterX{\kldivx}[2]{(}{)}{%
  #1\;\delimsize\|\;#2%
}
\newcommand{\kldiv}{D_{\text{KL}}\kldivx}

\cvprfinalcopy % *** Uncomment this line for the final submission

\def\cvprPaperID{3890} % *** Enter the CVPR Paper ID here

\ifcvprfinal\fi
\begin{document}

%%%%%%%%% TITLE
\title{Accurate and Diverse Sampling of Sequences based on a\\ ``Best of Many'' Sample Objective}

\author{Apratim Bhattacharyya, Bernt Schiele, Mario Fritz\\ 
Max Planck Institute for Informatics, Saarland Informatics Campus, Saarbr\"{u}cken, Germany \\
\texttt{\{abhattac, schiele, mfritz\}@mpi-inf.mpg.de}  }

\maketitle
%\thispagestyle{empty}

%%%%%%%%% ABSTRACT
\begin{abstract}
For autonomous agents to successfully operate in the real world, anticipation of future events and states of their environment is a key competence. This problem has been formalized as a sequence extrapolation problem, where a number of observations are used to predict the sequence into the future. Real-world scenarios demand a model of uncertainty of such predictions, as predictions become increasingly uncertain -- in particular on long time horizons. While impressive results have been shown on point estimates, scenarios that induce multi-modal distributions over future sequences remain challenging. Our work addresses these challenges in a Gaussian Latent Variable model for sequence prediction. Our core contribution is a ``Best of Many'' sample objective that leads to more accurate and more diverse predictions that better capture the true variations in real-world sequence data. Beyond our analysis of improved model fit, our models also empirically outperform prior work on three diverse tasks ranging from traffic scenes to weather data.
\end{abstract}

%%%%%%%%% BODY TEXT
\section{Introduction}

%We often encounter tasks in diverse fields ranging from autonomous driving to weather forecasting which require extrapolation into the future e.g. pedestrian trajectory prediction, steering angle prediction, precipitation forecasting. 
% Bernt: I rephrased the first sentence as it was quite long and not so easy to read
Predicting the future is important in many scenarios ranging from autonomous driving to precipitation forecasting. Many of these tasks can be formulated as sequence prediction problems. Given a past sequence of events, probable future outcomes are to be predicted. 

\begin{figure}[h]
\includegraphics[height=5.8cm]{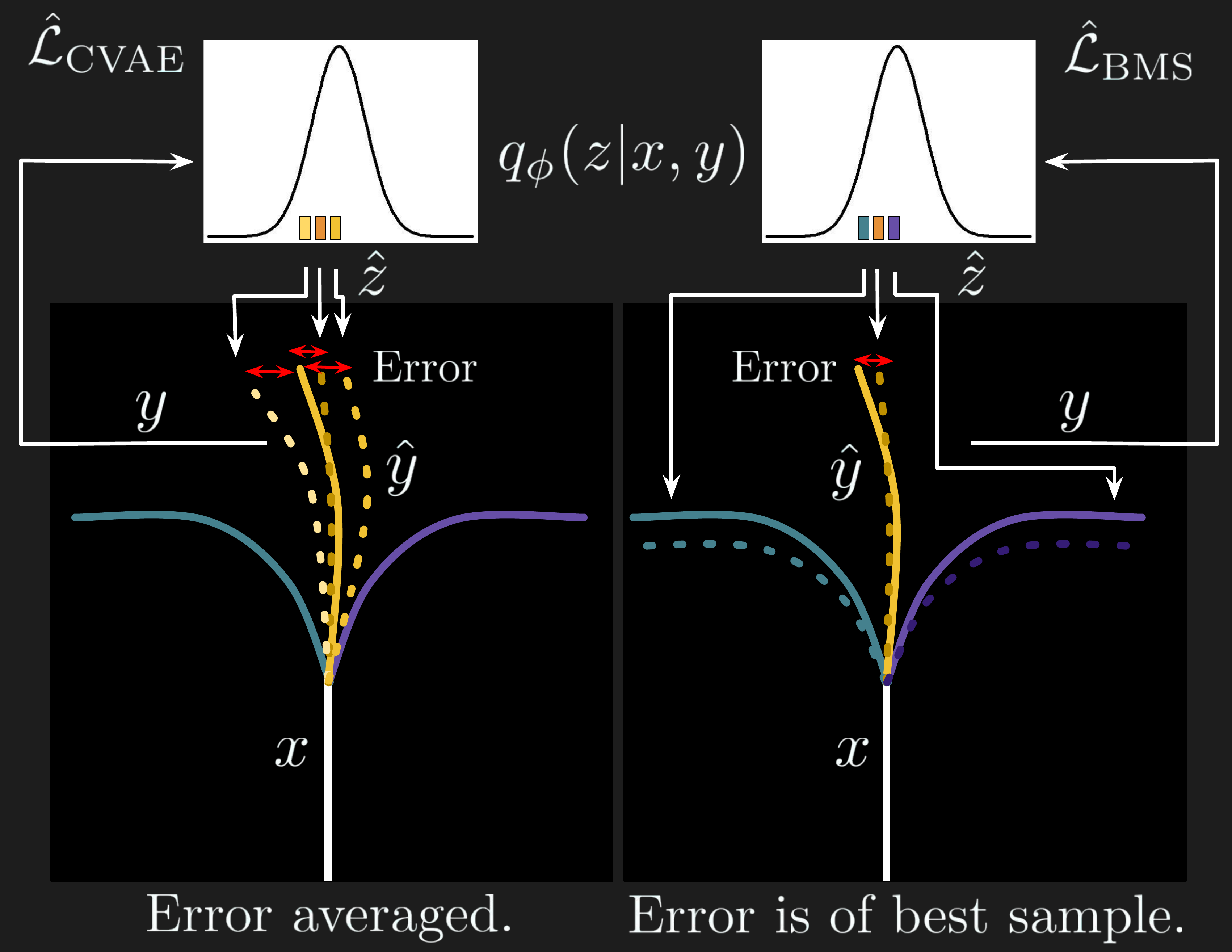}
\caption{Comparison between our ``Best of Many'' sample objective and the standard CVAE objective.}
  \label{fig:cgm}
\end{figure}

Recurrent Neural Networks (RNN) especially LSTM formulations are state-of-the-art models for sequence prediction tasks \cite{alahi2016social,xu2016end,finn2016unsupervised,xingjian2015convolutional}. These approaches predict only point estimates. However, many sequence prediction problems are only partially observed or stochastic in nature and hence the distribution of future sequences can be highly multi-modal. Consider the task of predicting future pedestrian trajectories. In many cases, we do not have any information about the intentions of the pedestrains in the scene. A pedestrian after walking over a Zerba crossing might decide to turn either left or right. A point estimate in such a situation would be highly unrealistic. Therefore, in order to incorporate uncertainty of future outcomes, we are interested in \emph{structured predictions}. Structured prediction in this context implies learning a one to many mapping of a given fixed sequence to plausible future sequences \cite{sohn2015learning}. This leads to more realistic predictions and enables probabilistic inference. 

Recent work \cite{lee2017desire} has proposed deep conditional generative models with Gaussian latent variables for structured sequence prediction. The Conditional Variational Auto-Encoder (CVAE) framework \cite{sohn2015learning} is used in \cite{lee2017desire} for learning of the Gaussian Latent Variables. We identify two key limitations of this CVAE framework. First, the currently used objectives hinder learning of diverse samples due to a marginalization over multi-modal futures. Second, a mismatch in latent variable distribution between training and testing leads to errors in model fitting. We overcome both challenges which results in more accurate and diverse  samples -- better capturing the true variations in data. 
%In addition, prior work \cite{lee2017desire} focuses only on vehicles and pedestrians in street scenes -- while the application domain is diverse.
Our main contributions are:
\begin{enumerate*}
\item We propose a novel ``best of many'' sample objective;
\item We analyze the benefits of our `best of many'' sample objective analytically as well as show an improved fit of latent variables on models trained with this novel objective compared to prior approaches;
\item We also show for the first time that this modeling paradigm extends to full-frame images sequences with diverse multi-modal futures; 
\item We demonstrate improved accuracy as well as diversity of the generated samples on three diverse tasks: MNIST stroke completion, Stanford Drone Dataset and HKO weather data. On all three datasets we consistently outperform the state of the art and baselines.
\end{enumerate*}

%-------------------------------------------------------------------------
\section{Related Work}
\myparagraph{Structured Output Prediction.} Stochastic feed-forward neural networks (SFNN) \cite{tang2013learning} model multi-modal conditional distributions through binary stochastic hidden variables. During training multiple samples are drawn and weighted according to importance-weights. However, due to the latent variables being binary SFNNs are hard to train on large datasets. There has been several efforts to make training more efficient for binary latent variables \cite{raiko2014techniques,gu2015muprop,mnih2016variational,lee2017simplified}. However, not all tasks can be efficiently modelled with binary hidden variables. In \cite{sohn2015learning}, Gaussian hidden variables are considered where the re-parameterization trick can be used for learning on large datasets using stochastic optimization. Inspired by this technique we model Gaussian hidden variables for structured sequence prediction tasks. 

\myparagraph{Variational Autoencoders.} Variational learning has enabled learning of deep directed graphical models (Conditional Generative Models) with Gaussian latent variables on large datasets \cite{kingma2013auto,kingma2014semi,jimenez2014stochastic}. Model training is made possible through stochastic optimization by the use of a variational lower bound of the data log-likelihood and the re-parameterization trick. In \cite{burda2015importance} a tighter lower bound on the data log-likelihood is introduced and multiple samples are used during training which are weighted according to importance weights. They show empirically that their IWAE framweork can learn richer latent space representations. However, these models do not consider conditional distributions for structured output prediction. Conditional variational auto-encoders (CVAE) \cite{sohn2015learning} extend the VAE framework of \cite{kingma2013auto} to model conditional distributions for structured output prediction by introducing the CVAE objective which maximizes a lower bound on the conditional data log liklihood. The CVAE framework has been used for a variety of tasks. Examples include, generation of likely future frames given a of a video \cite{visualdynamics16,fragkiadaki2017motion}, diverse images of clothed people conditioned on their silhouette~\cite{lassner2017generative}, and trajectories of basketball players using pictorial representations \cite{acunaunsupervised}. However, the gap between the training and test latent variable distributions cannot be fully closed by the CVAE objective function. We consider a new multi-sample objective which relaxes the constraints on the recognition network by encouraging diverse sample generation and thus leads to a better match between the training and test latent variable distributions.

In \cite{fragkiadaki2017motion} a multi-sample objective for Conditional Generative Models is also proposed in order to better capture multi-modal distributions. However, the key difference to our proposed objective is that during training samples are drawn from the prior. Sampling directly from the prior leads to high variance model updates. In contrast, we perform importance sampling through a  jointly learned proposal distribution in order to deal with this problem.

\myparagraph{Multiple Choice Learning.} In \cite{guzman2012multiple} a Multiple Choice Learning framework is presented where multiple models are learned to produce diverse predictions. During training, training examples are iteratively reassigned to the minimum loss model. The models are trained to convergence using these samples. Similarly in \cite{dey2015predicting} multiple diverse models are iteratively learned to solve submodular optimization tasks. As these frameworks require expensive retraining, in \cite{lee2016stochastic} a Stochastic Multiple Choice Learning framework is proposed. Here instead of iterative reassignment, for each example in a mini-batch, the error is back-propagated to only the model with the minimum loss (best) model. In contrast to our approach, these approaches require the learning of multiple models while our approach efficiently learns only one model without compromising on diversity of predictions.

\myparagraph{Recurrent Neural Networks.} Recurrent Neural Networks (RNNs) are state of the art methods for variety of sequence learning tasks \cite{graves2013generating,sutskever2014sequence}. In this work, we focus on sequence to sequence regression tasks, in particular, trajectory prediction and image sequence prediction. RNNs have been used for pedestrian trajectory prediction. In \cite{alahi2016social}, trajectories of multiple people in a scene are jointly modelled in a social context. However, even though the distribution of pedestrian trajectories are highly multimodal (with diverse futures), only one mean estimate is modelled. \cite{lee2017desire} jointly models multiple future pedestrian trajectories using a recurrent CVAE sampling module. Samples generated are refined and ranked using image and social context features. While our trajectory prediction model is similar to the sampling module of \cite{lee2017desire}, we focus on improving the sampling module by our novel multi-sample objective function. Convolutional RNNs \cite{xingjian2015convolutional} have been used for image sequence prediction. Examples include,  robotic arm movement prediction \cite{finn2016unsupervised} and precipitation now-casting \cite{xingjian2015convolutional,shi2017deep}. In this work, we extend the model of \cite{xingjian2015convolutional} for structured sequence prediction by conditioning predictions on Gaussian latent variables. Furthermore, we show that optimization using our novel multi-sample objective leads to improved results over the standard CVAE objective.

\section{Structured Sequence Prediction with Gaussian Latent Variables}

We begin with an overview of deep conditional generative models with gaussian latent variables and the CVAE framework with the corresponding objective \cite{sohn2015learning} used for training. Then, we introduce our novel ``best-of-many'' samples objective function. Thereafter, we introduce the conditional generative models which serve as the test bed for our novel objective. We first describe our model for structured trajectory prediction which is similar to the sampling module of \cite{lee2017desire} and consider extensions which additionally conditions on visual input and generates full image sequences. %We propose the first RNN based model for structure image sequence prediction. 
%Using these models, we show that our novel ``best-of-many'' samples objective leads to superior performance on both structured trajectory and image sequence prediction tasks.

\begin{figure}[h]
\includegraphics[height=2.5cm]{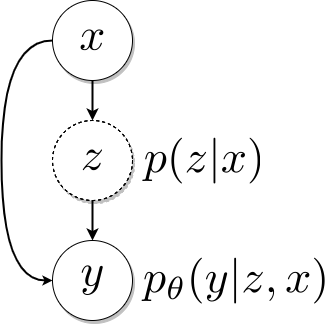}
\caption{Conditional generative models.}
  \label{fig:cgm}
\end{figure}

We consider deep conditional generative models of the form shown in \autoref{fig:cgm}. Given an input sequence $x$, a latent variable $\hat{z}$ is drawn from the conditional distribution $p(z | x)$ (assumed Gaussian). The output sequence $\hat{y}$ is then sampled from the distribution $p_{\theta}(y | x, z)$ of our conditional generative model with parameterized by $\theta$. The latent variables $z$ enables one-to-many mapping and the learning of multiple modes of the true posterior distribution $p(y | x)$. In practice, the simplifying assumption is made that $z$ is independent of $x$ and $p(z | x)$ is $\mathcal{N}(0,I)$.  Next, we discuss the training of such models.

\subsection{Conditional Variational Auto-encoder Based Training Objective}
We would like to maximize the data log-likelihood $p_{\theta}(y \mid x)$. To estimate the data log-likelihood of our model $p_{\theta}$, one possibility is to perform Monte-Carlo sampling of the latent variable $z$. For $T$ samples, this leads to the following estimate,
\begin{align}\label{eq2}
\hat{\mathcal{L}}_{\text{MC}} &= \log\Big( \frac{1}{T} \sum_{i=1}^{T} p_{\theta}(y | \hat{z}_{i}, x) \, \Big), \,\,\,  \hat{z}_{i} \sim \mathcal{N}(0,I).
\end{align}
This estimate is unbiased but has high variance \cite{mnih2016variational}. We would underestimate the log-likelihood for some samples and overestimate for others, especially if $T$ is small. This would in turn lead to high variance weight updates.

We can reduce the variance of updates by estimating the log-likelihood through importance sampling during training. As described in \cite{sohn2015learning}, we can sample the latent variables $z$ from a recognition network $q_{\phi}$ using the re-parameterization trick \cite{kingma2013auto}. The data log-likelihood is,
\begin{align}\label{eq3}
\begin{split}
&\log(p_{\theta}(y \mid x)) = \\
 & \log\Big( \int p_{\theta}(y | z, x) \, \frac{p(z | x)}{q_{\phi}(z | x, y)} \, q_{\phi}(z | x, y) \, dz \, \Big).
\end{split}
\end{align}
The integral in (\ref{eq3}) is computationally intractable. In \cite{sohn2015learning}, a variational lower bound of the data log-likelihood (\ref{eq3}) is derived, which can be estimated empirically using Monte-Carlo integration (also used in \cite{lee2017desire}),
\begin{align}\label{eq4}
\begin{split}
&\hat{\mathcal{L}}_{\text{CVAE}} =\frac{1}{T} \sum_{i=1}^{T} \log \,p_{\theta}(y | \hat{z}_{i}, x) \\
&- \kldiv{q_{\phi}(z | x, y)}{p(z | x)}, \,\,  \hat{z}_{i} \sim q_{\phi}(z | x, y).
\end{split}
\end{align}
The lower bound in (\ref{eq4}) weights all samples ($\hat{z}_{i}$) equally and so they must all ascribe high probability to the data point $(x,y)$. This introduces a strong constraint on the recognition network $q_{\phi}$. Therefore, the model is forced to trade-off between a good estimate of the data log-likelihood and the KL divergence between the training and test latent variable distributions. One possibility to close the gap introduced between the training and test pipelines, as described in \cite{sohn2015learning}, is to use an hybrid objective of the form $(1 - \alpha) \hat{\mathcal{L}}_{\text{MC}} + \alpha \, \hat{\mathcal{L}}_{\text{CVAE}}$. Although such an hybrid objective has shown modest improvement in performance in certain cases, we could not observe any significant improvement over the standard CVAE objective in our structured sequence prediction tasks. In the following, we derive our novel ``best-of-many-samples'' objective which on the one hand encourages sample diversity and on the other hand aims to close the gap between the training and testing pipelines.

\begin{figure*}[t]
\centering
\hfill
\begin{minipage}{.245\textheight}
  \centering
  \includegraphics[width=1\linewidth]{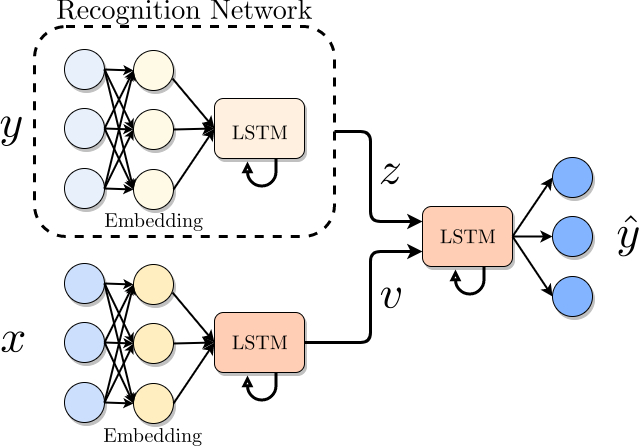}
  \subcaption{Our model for structured trajectory prediction.}
  \label{fig:traj_model}
\end{minipage}%
\hfill
\begin{minipage}{.345\textheight}
  \centering
  \includegraphics[width=1\linewidth]{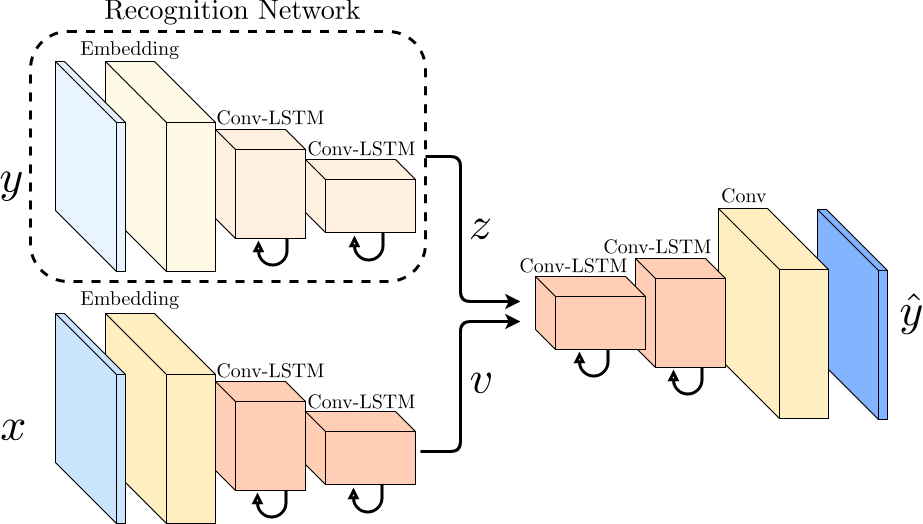}
  \subcaption{Our model for structured image sequence prediction.}
  \label{fig:imseq_model}
\end{minipage}
\caption{Our model architectures. The recognition networks are only available during training.}
\end{figure*}

\subsection{Best of Many Samples Objective}
Here, we propose our objective which unlike (\ref{eq4}) does not weight each sample equally. Consider the functions $f_{1}(z) = \nicefrac{p(z | x)}{q_{\phi}(z | x, y)}$ and $f_{2}(z) = p_{\theta}(y | z, x) \times q_{\phi}(z | x, y)$ in (\ref{eq3}). We cannot evaluate $f_{2}(z)$ directly for Monte-Carlo samples. Notice, however that both $f_{1}(z)$ and $f_{2}(z)$ are continuous and positive. As $q_{\theta}(z | x, y)$ is normally distributed, the integral above can be very well approximated on a large enough bounded interval $[a,b]$. Therefore, we can use the First Mean Value Theorem of Integration \cite{comenetz2002calculus}, to separate the functions $f_{1}(z)$ and $f_{2}(z)$ in (\ref{eq3}),
\begin{align}\label{eq5}
\begin{split}
\log(p_{\theta}(y | x)) &= \log\Big( \int_{a}^{b} p_{\theta}(y | z, x) \, q_{\phi}(z | x, y) \, dz \, \Big) \\
&+ \log\Big( \frac{p(z^{\prime} | x)}{q_{\phi}(z^{\prime} | x, y)} \Big), \,\, z^{\prime} \in (a,b).
\end{split}
\end{align}
We can lower bound (\ref{eq5}) with the minimum of the term on the right,
\begin{align}\label{eq6}
\begin{split}
\log(p_{\theta}(y | x)) &\geq \log\Big( \int_{a}^{b} p_{\theta}(y | z, x) \, q_{\phi}(z | x, y) \, dz \, \Big) \\
&+ \min_{z^{\prime} \in (a,b)}\Big( \log\Big( \frac{p(z^{\prime} | x)}{q_{\phi}(z^{\prime} | x, y)} \Big)\Big)
\end{split}
\end{align}
We can estimate the first term on the right of (\ref{eq6}) using Monte-Carlo integration. The minimum in the second term on the right of (\ref{eq6}) is difficult to estimate, therefore we approximate it by the KL divergence over the full distribution. The KL divergence heavily penalizes $q_{\phi}(z | x, y)$ when it is high for low values $p(z | x)$ (which leads to low value of the ratio of the distributions). This leads to the following ``many-sample'' objective, (more details in the supplementary section),
\begin{align}\label{eq7}
\begin{split}
&\hat{\mathcal{L}}_{\text{MS}} = \log\Big( \frac{1}{T} \sum_{i=1}^{T} p_{\theta}(y | \hat{z}_{i}, x) \, \Big) \\
&- \kldiv{q_{\phi}(z | x, y)}{p(z | x)}, \,\,  \hat{z}_{i} \sim q_{\phi}(z | x, y).
\end{split}
\end{align}
Compared to the CVAE objective (\ref{eq3}), the recognition network $q_{\phi}$ now has multiple chances to draw samples with high posterior probability ($p_{\theta}(y \mid z, x)$). This encourages diversity in the generated samples. Furthermore, the data log-likelihood (\ref{eq3}) estimate in this objective is tighter as $\hat{\mathcal{L}}_{\text{MS}} \geq \hat{\mathcal{L}}_{\text{CVAE}}$ follows from the Jensen's inequality. Therefore, this bound loosens the constrains on the recognition network $q_{\phi}$ and allows it more closely match the latent variable distribution $p(z | x)$. However, as we focus on regression tasks, probabilities are of the form $e^{-\text{MSE}(\hat{y},y)}$. Therefore, in practice the Log-Average term can cause numerical instabilities due to limited machine precision in representing the probability  $e^{-\text{MSE}(\hat{y},y)}$. Therefore, we use a ``Best of Many Samples'' approximation $\hat{\mathcal{L}}_{\text{BMS}}$ of (\ref{eq7}). We can pull the constant $\nicefrac{1}{T}$ term outside the average in (\ref{eq7}) and approximate the sum with the maximum,
\begin{align}\label{eq8}
\begin{split}
&\hat{\mathcal{L}}_{\text{MS}} = \log\Big( \sum_{i=1}^{T} p_{\theta}(y | \hat{z}_{i}, x) \, \Big) - \log(T) \\
&- \kldiv{q_{\phi}(z | x, y)}{p(z | x)}, \,\,  \hat{z}_{i} \sim q_{\phi}(z | x, y)
\end{split} 
\end{align}
\begin{align}\label{eq9}
\begin{split}
\hat{\mathcal{L}}_{\text{MS}} &\geq \hat{\mathcal{L}}_{\text{BMS}} = \max_{i} \big( \log( p_{\theta}(y | \hat{z}_{i}, x) ) \, \big) - \log(T) \\
&-\kldiv{q_{\phi}(z | x, y)}{p(z | x)}, \,\,  \hat{z}_{i} \sim q_{\phi}(z | x, y). 
\end{split}
\end{align}

Similar to (\ref{eq7}), this objective encourages diversity and loosens the constrains on the recognition network $q_{\phi}$ as only the best sample is considered. During training, initially $p_{\theta}$ assigns low probability to the data for all samples $\hat{z}_{i}$. The $\log(T)$ difference between (\ref{eq7}) and (\ref{eq9}) would be dominated by the low data log-likelihood. Later on, as both objectives promote diversity, the Log-Average term in (\ref{eq7}) would be dominated by one term in the average. Therefore, (\ref{eq7}) would be well approximated by the maximum of the terms in the average.  Furthermore, (\ref{eq9}) avoids numerical stability issues.

%As we focus on regression tasks, the probability term $p_{\theta}(y \mid \hat{z}_{i}, x)$ takes the form $e^{-\text{MSE}(y,\hat{y})}$, where MSE is the Mean Squared Error. Therefore, the first term on the left of (\ref{eq8}) is of the form Log-Sum-Exp, $\log(e^{-x_{1}} + e^{-x_{2}} + .. + e^{-x_{n}})$ and is well approximated by $\max\left\{-x_{1},..,-x_{n}\right\}$ as in (\ref{eq9}).
%This allows for closer match between $q_{\phi}$ and the latent variable distribution $p(z \mid x)$ thereby closing the gap between the training and testing pipelines.

\begin{figure*}[t]
  
  \centering
  \resizebox{\textwidth}{!}{\begin{tabular}{ *7{c} }
    \cline{1-4} \cline{5-7}
     \multicolumn{1}{|c}{} & & & \multicolumn{1}{c|}{} & \multicolumn{1}{|c}{} & & \multicolumn{1}{c|}{} \\
    \multicolumn{1}{|c}{\includegraphics[height=2.35cm,trim={4cm 4cm 4cm 4cm},clip]{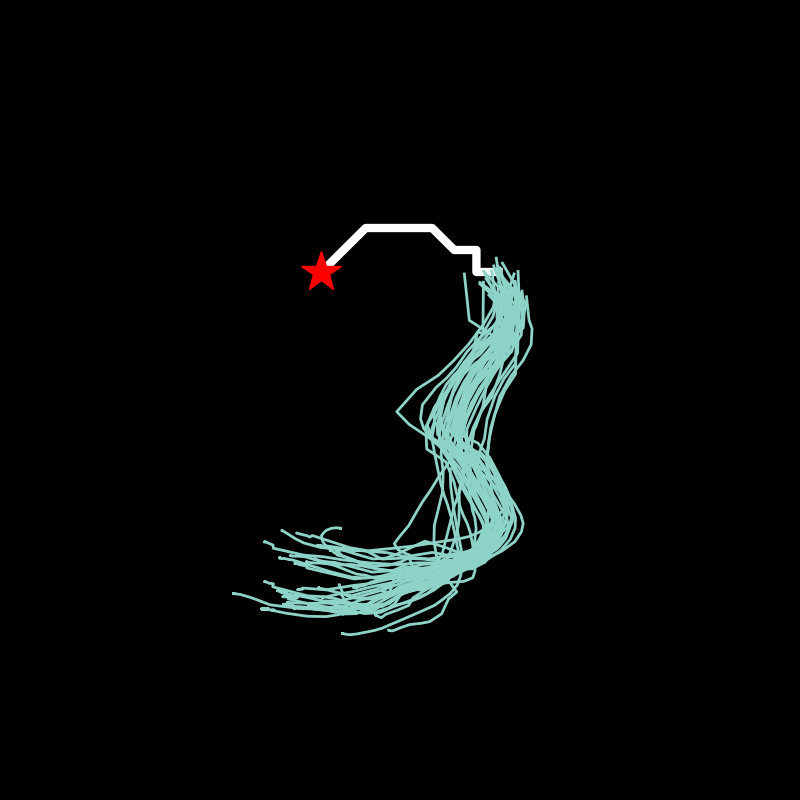}} &
    \includegraphics[height=2.35cm,trim={4cm 4cm 4cm 4cm},clip]{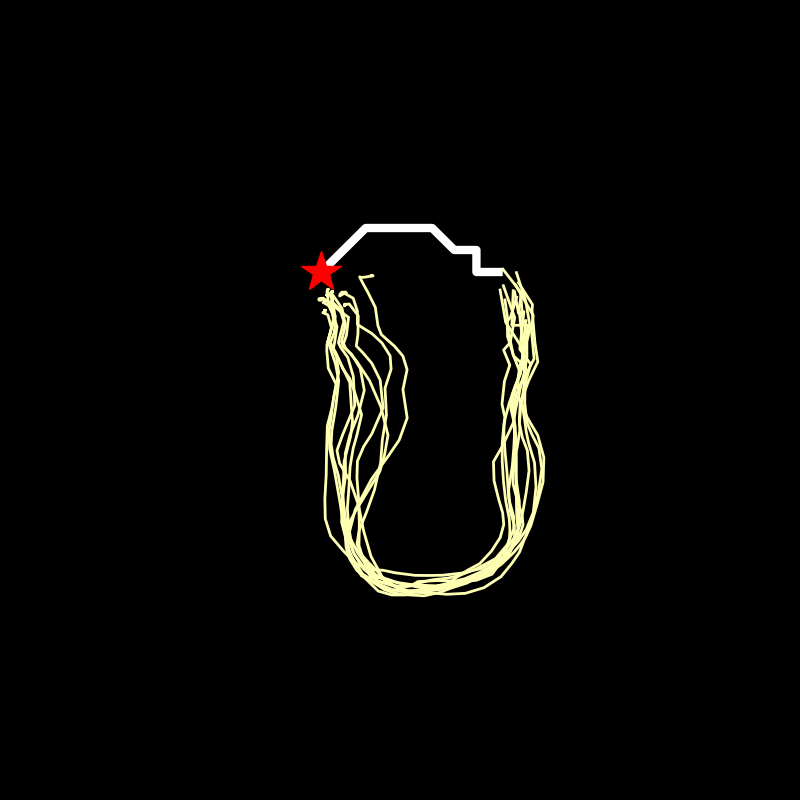} &
    \includegraphics[height=2.35cm,trim={4cm 4cm 4cm 4cm},clip]{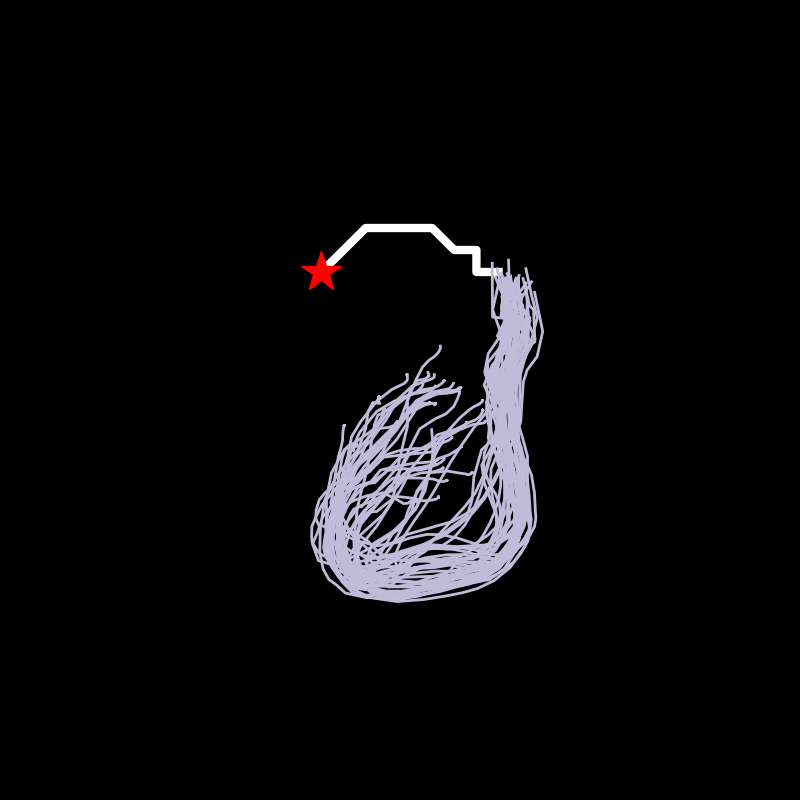} &
    \multicolumn{1}{c|}{\includegraphics[height=2.35cm,trim={4cm 4cm 4cm 4cm},clip]{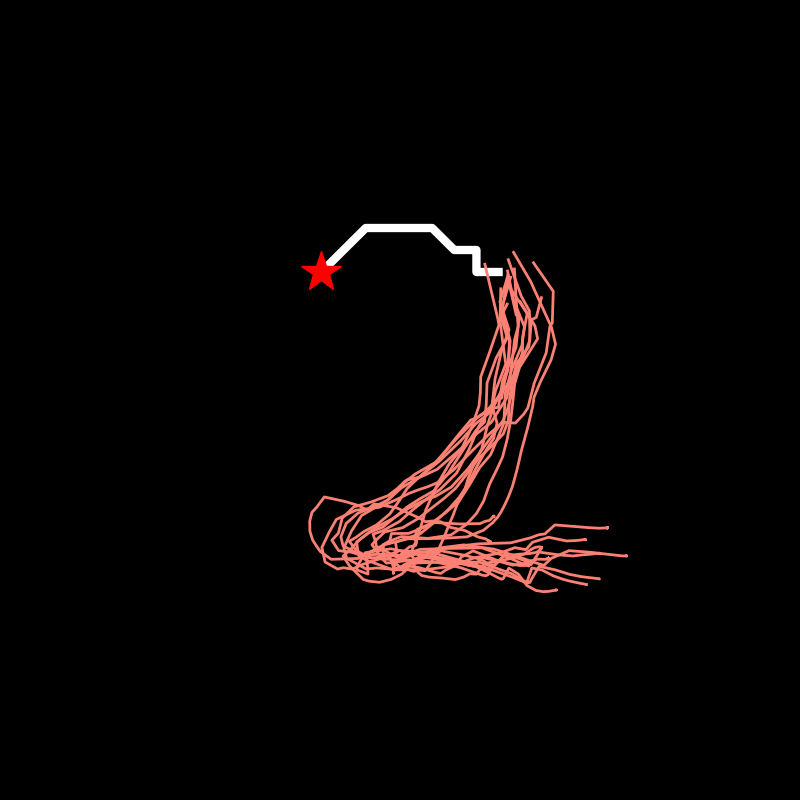}} &
    \multicolumn{1}{|c}{\includegraphics[height=2.35cm,trim={4cm 4cm 4cm 4cm},clip]{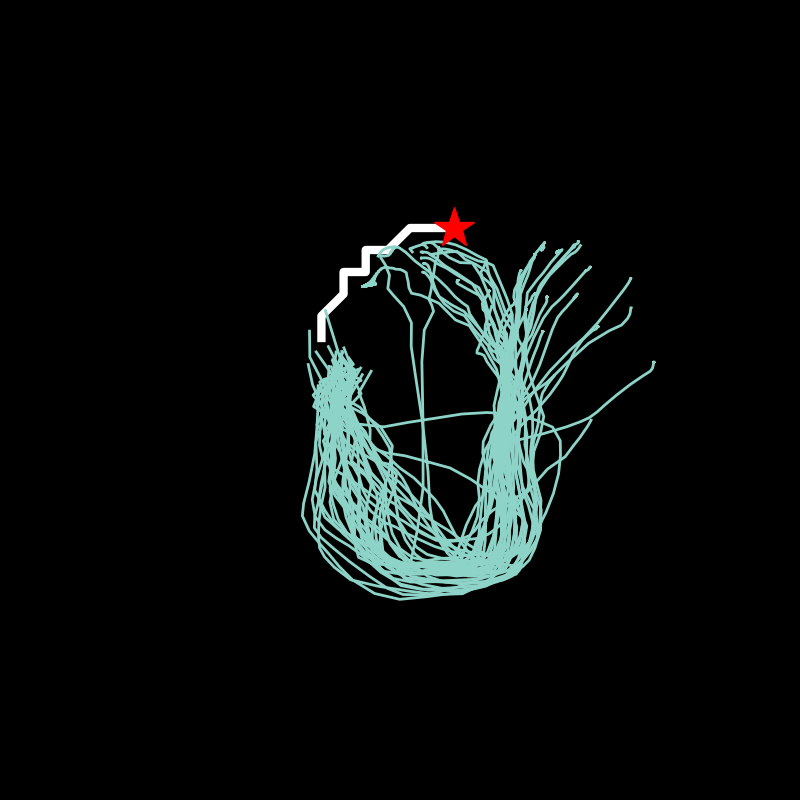}} &
    \includegraphics[height=2.35cm,trim={4cm 4cm 4cm 4cm},clip]{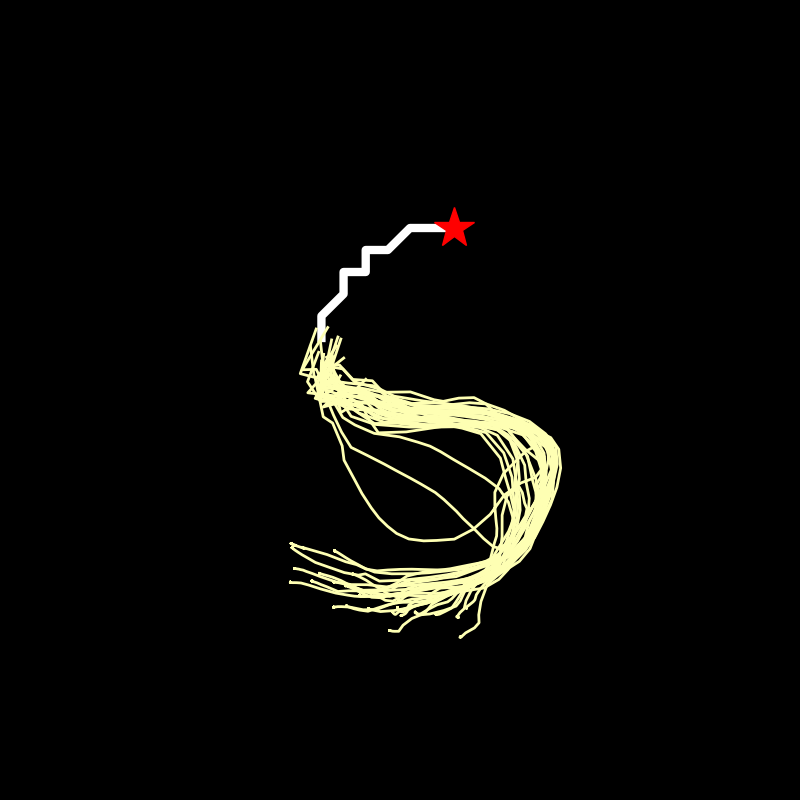} &
    \multicolumn{1}{c|}{\includegraphics[height=2.35cm,trim={4cm 4cm 4cm 4cm},clip]{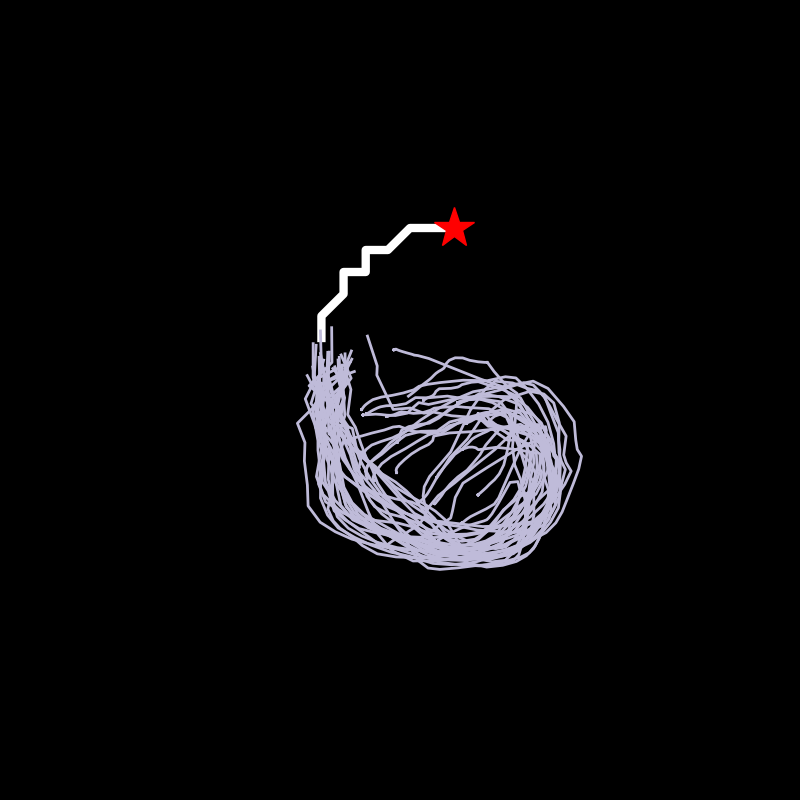}} \\
    
    \cline{1-4} \cline{5-7}
    
    & & & & & &  \\
    
    \cline{1-3} \cline{4-5} \cline{6-7}
    
    \multicolumn{1}{|c}{} & & \multicolumn{1}{c|}{} & \multicolumn{1}{|c}{} & \multicolumn{1}{c|}{} & \multicolumn{1}{|c}{} & \multicolumn{1}{c|}{} \\
    
    \multicolumn{1}{|c}{\includegraphics[height=2.35cm,trim={4cm 4cm 4cm 4cm},clip]{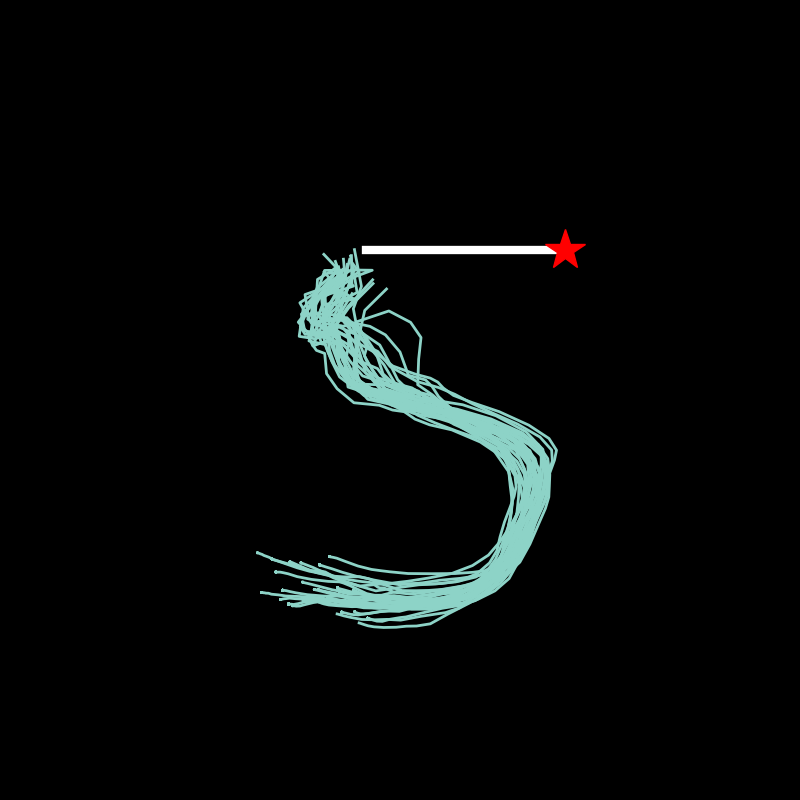}} &
    \includegraphics[height=2.35cm,trim={4cm 4cm 4cm 4cm},clip]{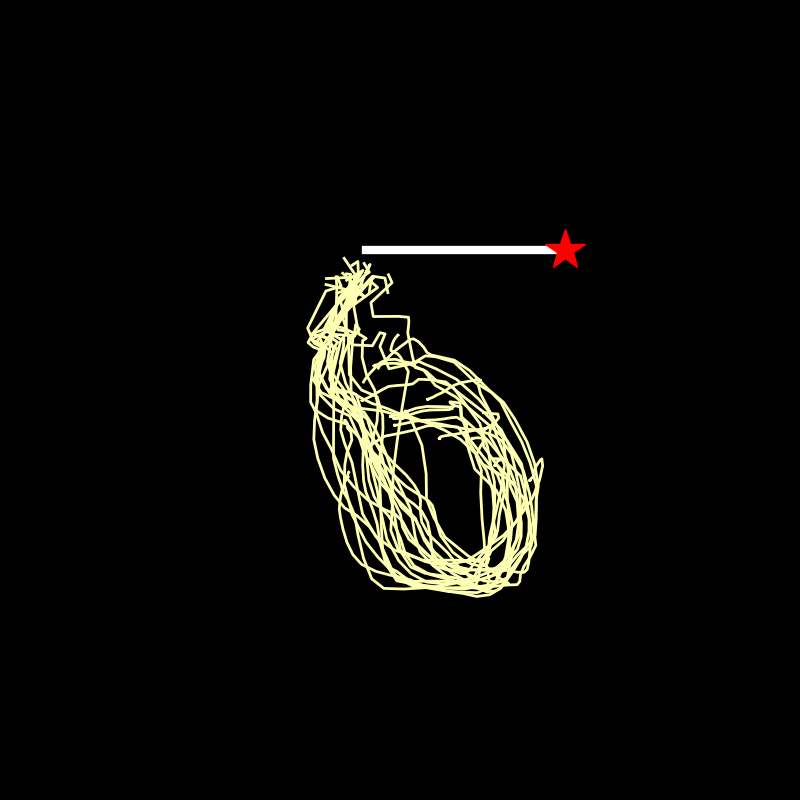} &
    \multicolumn{1}{c|}{\includegraphics[height=2.35cm,trim={4cm 4cm 4cm 4cm},clip]{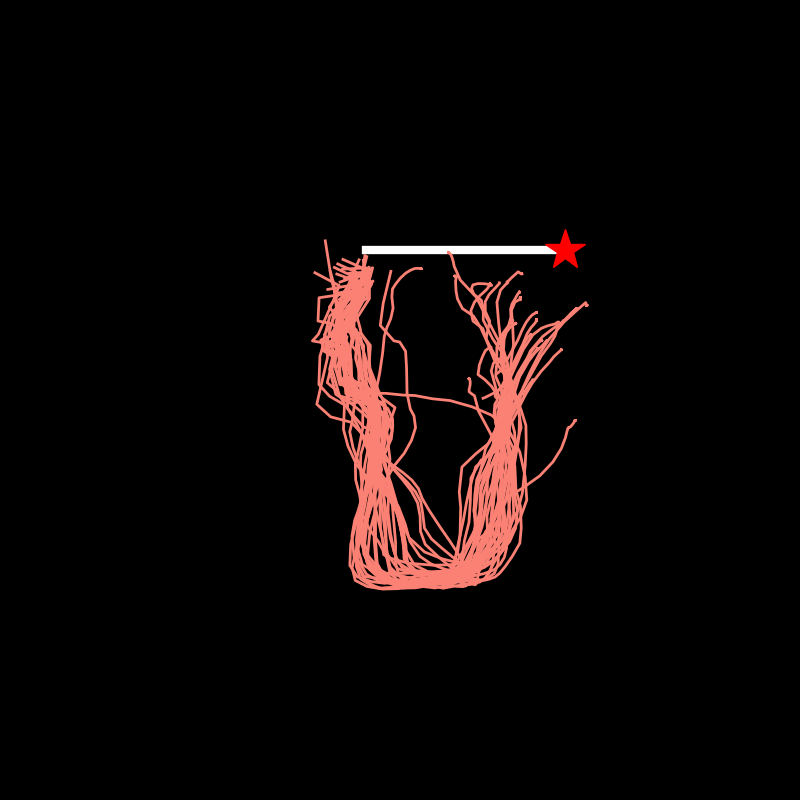}} &
    \multicolumn{1}{|c}{\includegraphics[height=2.35cm,trim={4cm 4cm 4cm 4cm},clip]{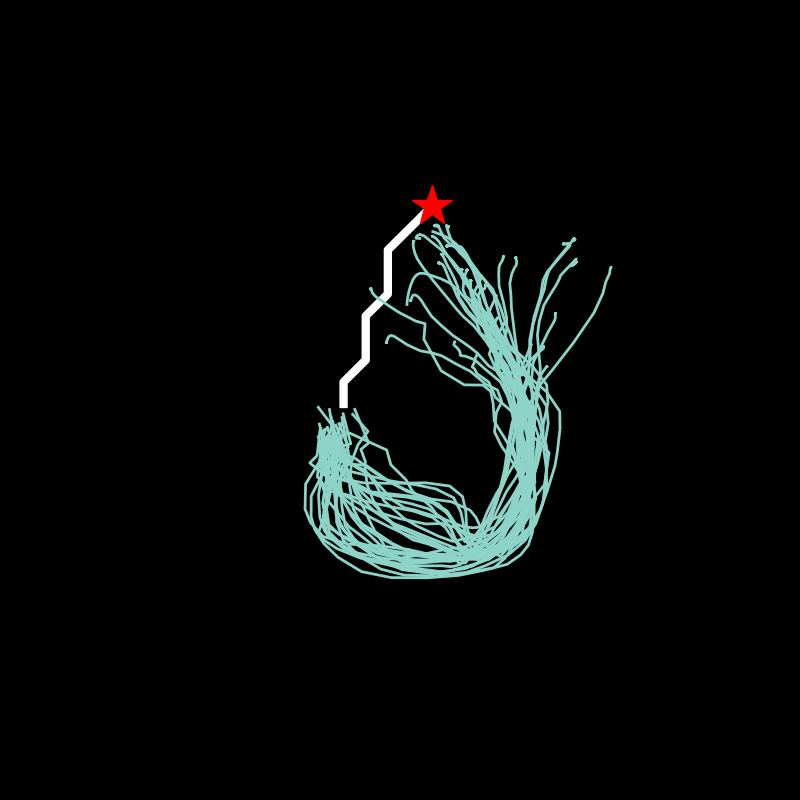}} &
    \multicolumn{1}{c|}{\includegraphics[height=2.35cm,trim={4cm 4cm 4cm 4cm},clip]{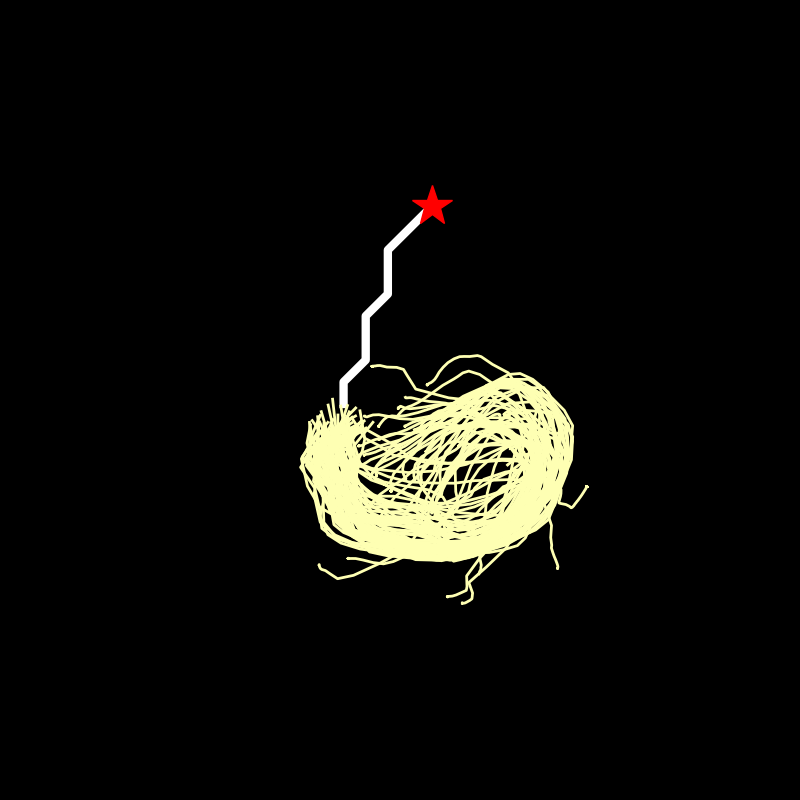}} &
    \multicolumn{1}{|c}{\includegraphics[height=2.35cm,trim={4cm 4cm 4cm 4cm},clip]{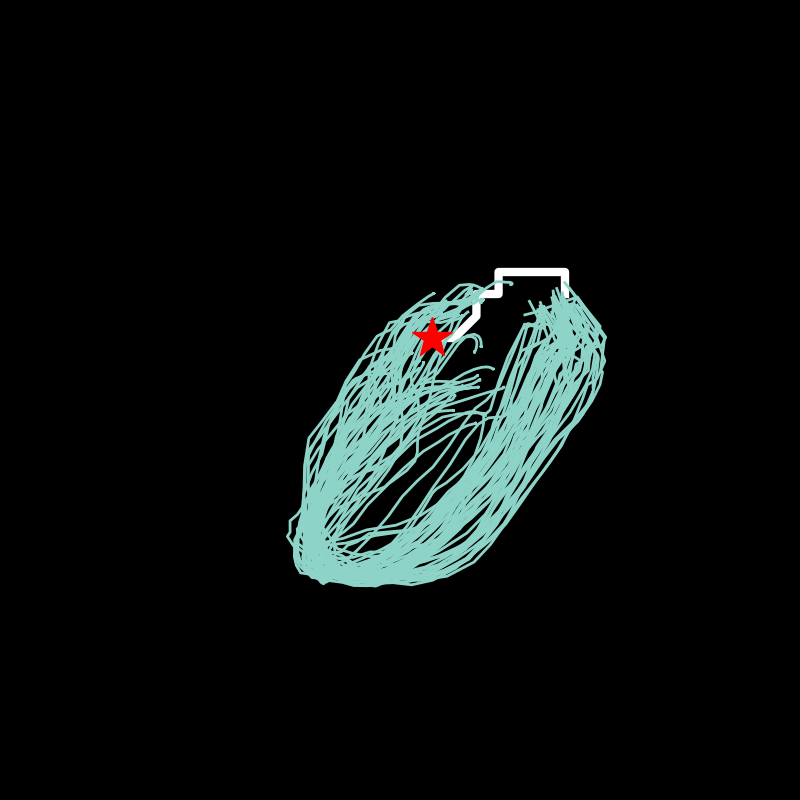}} &
    \multicolumn{1}{c|}{\includegraphics[height=2.35cm,trim={4cm 4cm 4cm 4cm},clip]{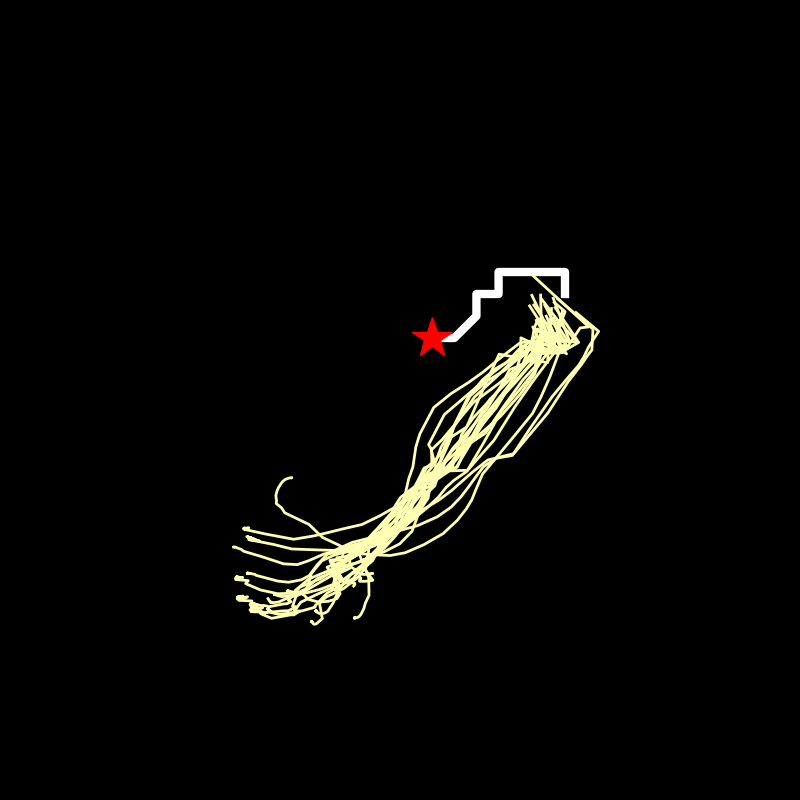}} \\
    
    \cline{1-3} \cline{4-5} \cline{6-7}
    
    \end{tabular}}
  \caption{Diverse samples drawn from our LSTM-BMS model trained using the $\hat{\mathcal{L}}_{\text{BMS}}$ objective, clustered using k-means. The number of clusters is set manually to the number of expected digits based on the initial stroke.}
   
  \label{fig:mnist_ex}
\end{figure*}

\begin{figure*}[h]
  
  \centering
  \begin{tabular}{ c@{\hskip 0.1cm}c@{\hskip 0.1cm}c@{\hskip 0.1cm}c@{\hskip 0.1cm}c@{\hskip 0.1cm}c@{\hskip 0.1cm}c }
    
    \includegraphics[height=2.35cm,trim={4cm 2cm 4cm 6cm},clip]{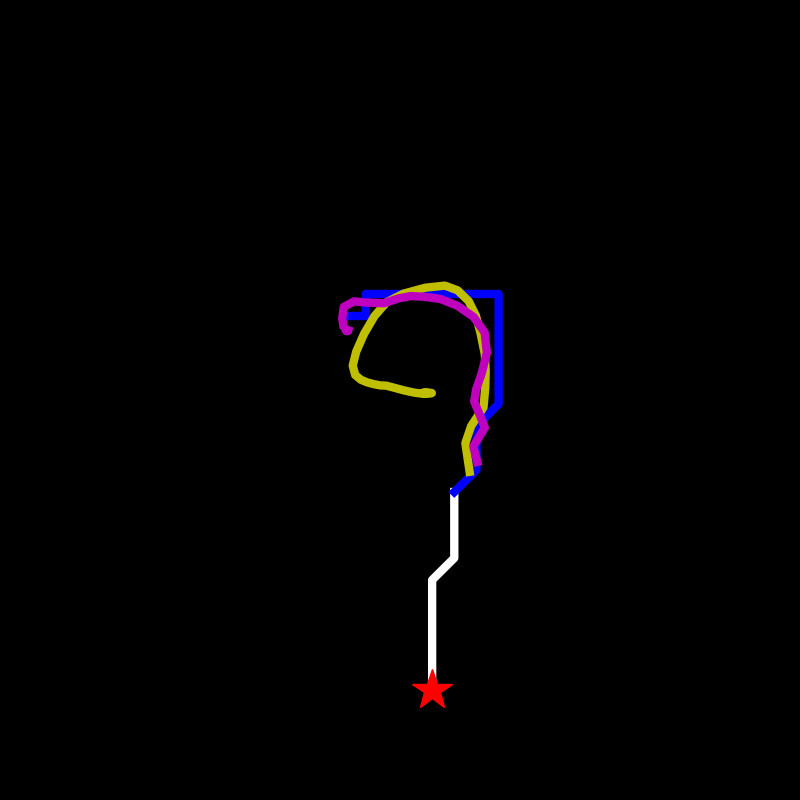} &
    \includegraphics[height=2.35cm,trim={4cm 4cm 4cm 4cm},clip]{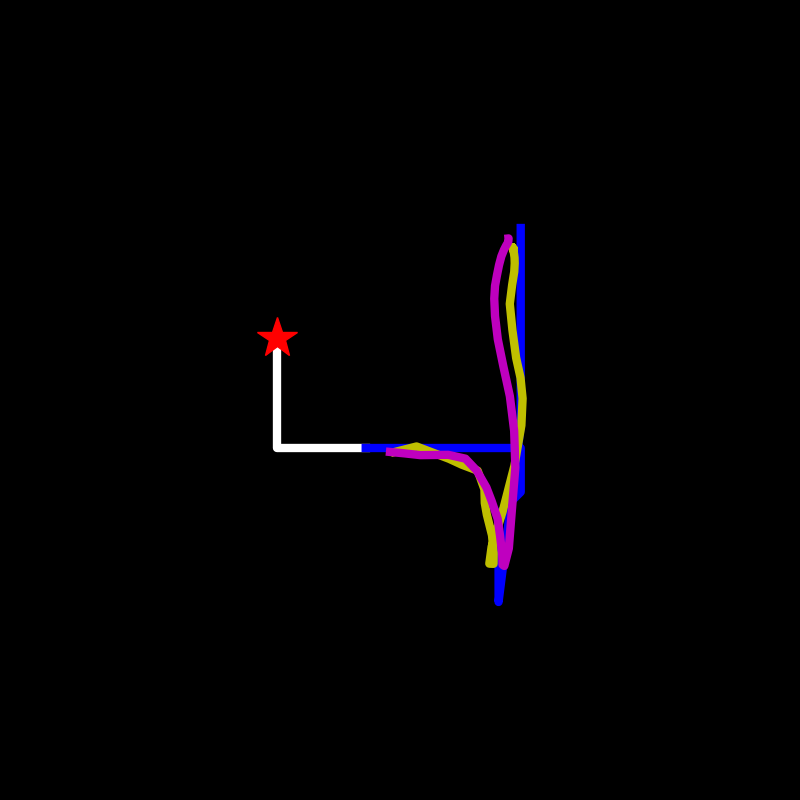} &
    \includegraphics[height=2.35cm,trim={4cm 4cm 4cm 4cm},clip]{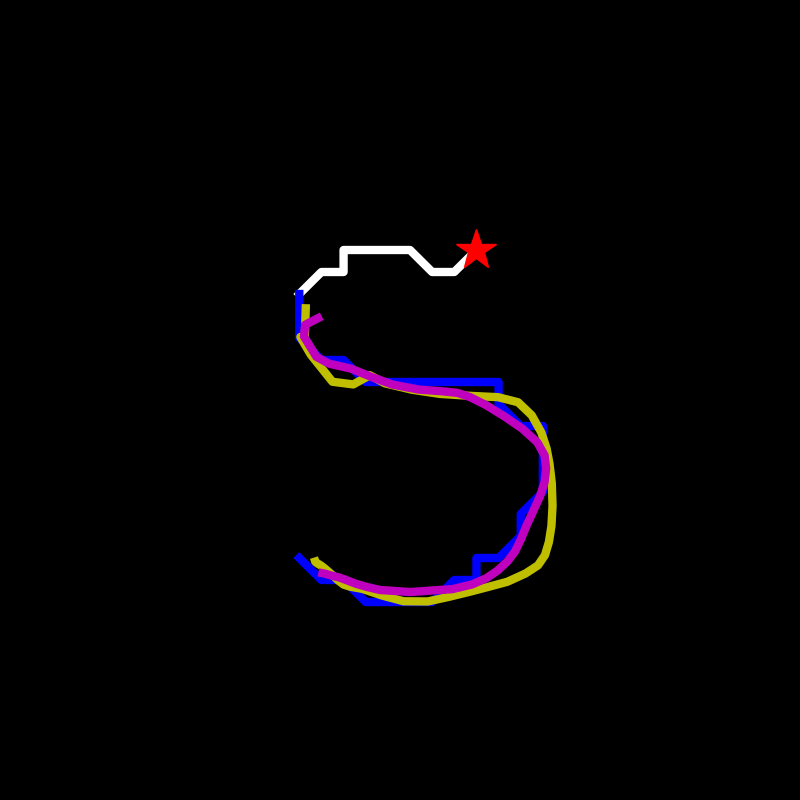} &
    \includegraphics[height=2.35cm,trim={4cm 3cm 4cm 5cm},clip]{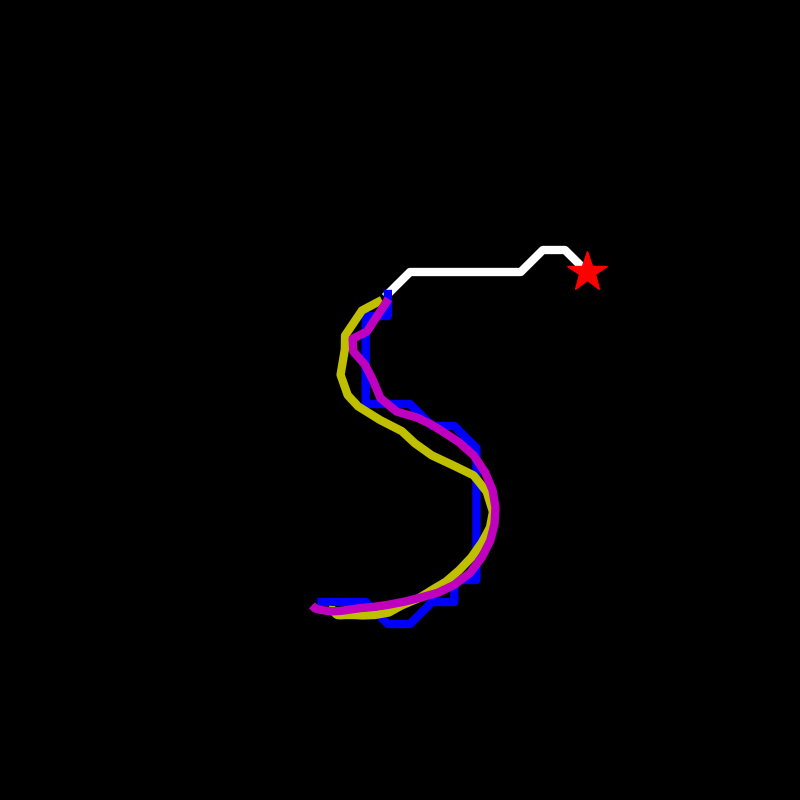} &
    \includegraphics[height=2.35cm,trim={4cm 5cm 4cm 3cm},clip]{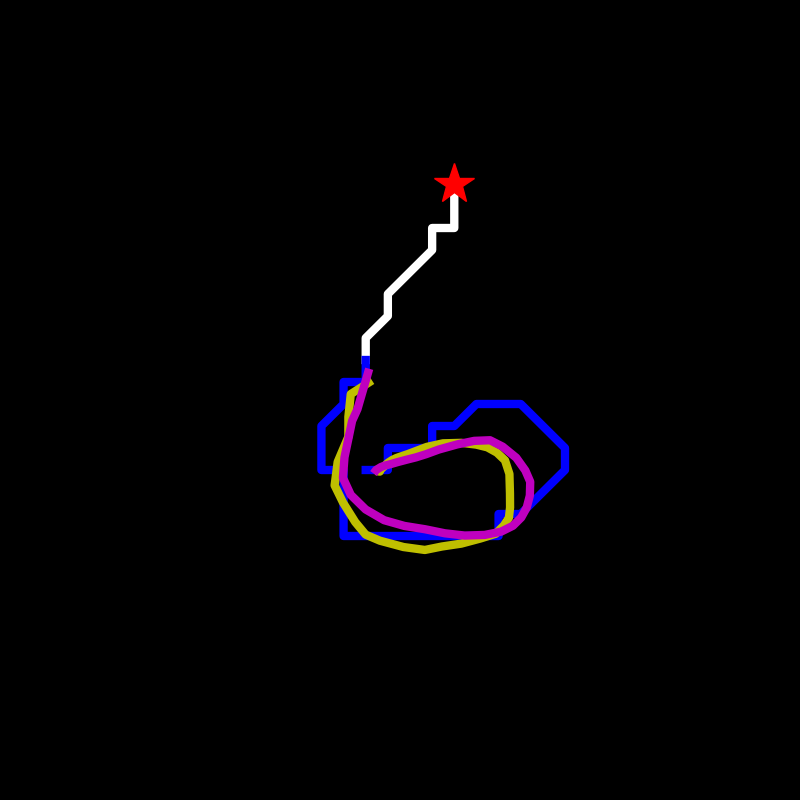} &
    \includegraphics[height=2.35cm,trim={4cm 4cm 4cm 4cm},clip]{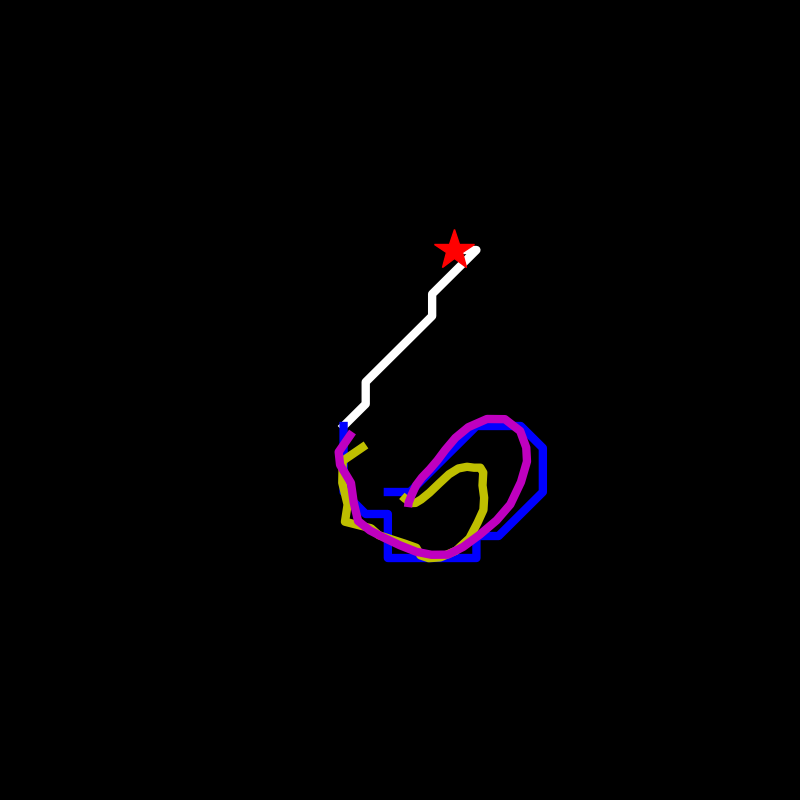} &
    \includegraphics[height=2.35cm,trim={4cm 2cm 4cm 6cm},clip]{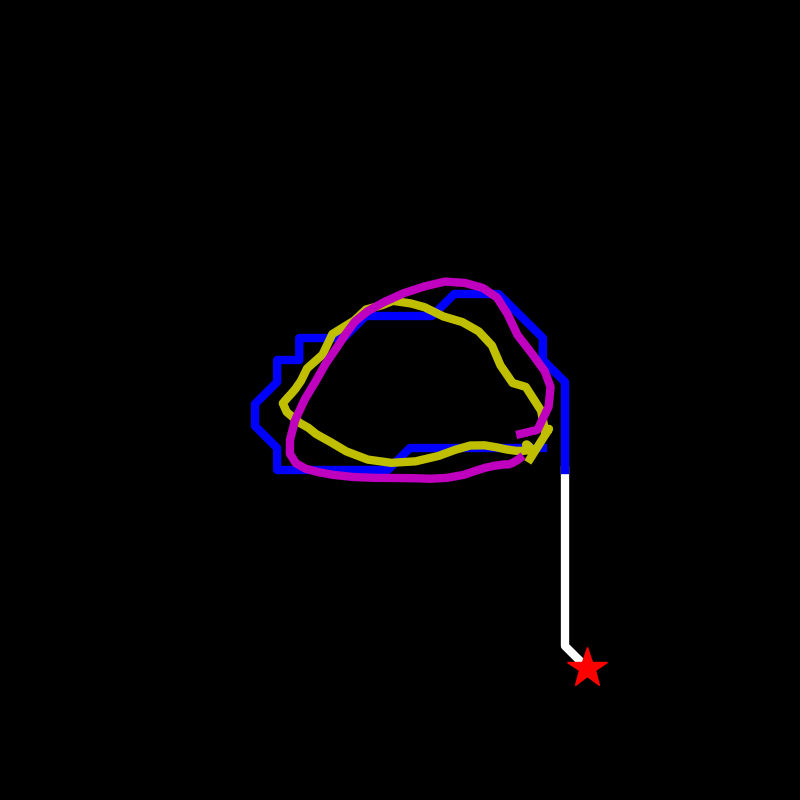} \\
    
    \end{tabular}
  \caption{Top 10\% of samples drawn from the LSTM-BMS model (\textcolor{black}{magenta}) and the LSTM-CVAE model (\textcolor{black}{yellow}), with the groundtruth in (\textcolor{black}{blue}).}
   
  \label{fig:mnist_comp}
\end{figure*}

\subsection{Model Architectures for Structured Sequence Prediction}\label{ssec:models}
We base our model architectures on RNN Encoder-Decoders. We use LSTM formulations as RNNs for structured trajectory prediction tasks (\autoref{fig:traj_model}) and Convolutional LSTM formulations (\autoref{fig:imseq_model}) for structured image sequence prediction tasks. During training, we consider LSTM recognition networks in case of trajectory prediction (\autoref{fig:traj_model}) and for image sequence prediction, we consider Conv-LSTM recognition networks (\autoref{fig:imseq_model}). Note that, as we make the simplifying assumption that $z$ is independent of $x$, the recognition networks are conditioned only on $y$.

\myparagraph{Model for Structured Trajectory Prediction.} Our model for structured trajectory prediction (see \autoref{fig:traj_model}) is similar to the sampling module of \cite{lee2017desire}. The input sequence $x$ is processed using an embedding layer to extract features and the embedded sequence is read by the encoder LSTM. The encoder LSTM produces a summary vector $v$, which is its internal state after reading the input sequence $x$. The decoder LSTM is conditioned on the summary vector $v$ and additionally a sample of the latent variable $z$. The decoder LSTM is unrolled in time and a prediction is generated by a linear transformation of it's output. Therefore, the predicted sequence at a certain time-step $\hat{y}^{t}$ is conditioned on the output at the previous time-step, the summary vector $v$ and the latent variable $z$. As the summary $v$ is deterministic given $x$, we have,

\begin{align*}
\begin{split}
\log(p_{\theta}(y | x)) &=  \sum_{t} \log(p_{\theta}(y^{t+1} | y^{t}, v) \, p(v | x))\\
  &=  \sum_{t} \log(p_{\theta}(y^{t+1} | y^{t}, x))\\
  &=  \int \sum_{t} \log(p_{\theta}(y^{t+1} | y^{t}, z, x) \, p_{\theta}(z | x)) \, dz.
\end{split}
\end{align*}

Conditioning the predicted sequence at all time-steps upon a single sample of $z$ enables $z$ to capture global characteristics (e.g. speed and direction of motion) of the future sequence and generation of temporally consistent sample sequences $\hat{y}$. 

\myparagraph{Extension with Visual Input.} In case of dynamic agents e.g. pedestrians in traffic scenes, the future trajectory is highly dependent upon the environment e.g. layout of the streets. Therefore, additionally conditioning samples on sensory input (e.g. visuals of the environment) would enable more accurate sample generation. We use a CNN to extract a summary of a visual observation of a scene. This visual summary is given as input to the decoder LSTM, ensuring that the generated samples are additionally conditioned on the visual input.

\myparagraph{Model for Structured Image Sequence Prediction.} If the sequence $(x,y)$ in question consists of images e.g. frames of a video, the trajectory prediction model \autoref{fig:traj_model} cannot exploit the spatial structure of the image sequence. More specifically, consider a pixel $y^{t+1}_{i,j}$ at time-step $t+1$ of the image sequence $y$. The pixel value at time-step $t+1$ depends upon only the pixel $y^{t}_{i,j}$ and a certain neighbourhood around it. Furthermore, spatially neighbouring pixels are correlated. This spatial structure can be exploited by using Convolutional LSTMs \cite{xingjian2015convolutional} as RNN encoder-decoders. Conv-LSTMs retain spatial information by considering the hidden states $h$ and cell states $c$ as 3D tensors -- the cell and hidden states are composed of vectors $c^{t}_{i,j}$, $h^{t}_{i,j}$ corresponding to each spatial position. New cell states, hidden states and outputs are computed using convolutional operations. Therefore, new cell states  $c^{t+1}_{i,j}$, hidden states $h^{t+1}_{i,j}$ depend upon only a local spatial neighbourhood of $c^{t}_{i,j}$, $h^{t}_{i,j}$, thus preserving spatial information.

We propose conditional generative models networks with Conv-LSTMs for structured image sequence prediction (\autoref{fig:imseq_model}). The encoder and decoder consists of two stacked Conv-LSTMs for feature aggregation. As before, the output is conditioned on a latent variable $z$ to model multiple modes of the conditional distribution $p(y \mid x)$. The future states of neighboring pixels are highly correlated. However, spatially distant parts of the image sequences can evolve independently. To take into account the spatial structure of images, we consider latent variables $z$ which are 3D tensors. As detailed in \autoref{fig:imseq_model}, the input image sequence $x$ is processed using a convolutional embedding layer. The Conv-LSTM reads the embedded input sequence and produces a 3D tensor $v$ as the summary. The 3D summary $v$ and latent variable $z$ is given as input to the Conv-LSTM decoder at every time-step. The cell state, hidden state or output at a certain spatial position, $c^{t}_{i,j}$, $h^{t}_{i,j}$, $y^{t}_{i,j}$, it is conditioned on a sub-tensor $\underline{z}_{i,j}$ of the latent tensor $z$. Spatially neighbouring cell states, hidden states (and thus  outputs) are therefore conditioned on spatially neighbouring sub-tensors $\underline{z}_{i,j}$. This coupled with the spatial information preserving property of Conv-LSTMs detailed above, enables $z$ to capture spatial location specific characteristics of the future image sequence and allows for modeling the correlation of future states of spatially neighboring pixels. This ensures spatial consistency of sampled output sequences $\hat{y}$. Furthermore, as in the fully connected case, conditioning the full output sequence sample $\hat{y}$ is on a single sample of $z$ ensures temporal consistency. 

\begin{table}[h]
  \begin{tabular}{cc}
    \toprule
    Method & CLL\\
    \midrule
    LSTM & 136.12 \\
    LSTM-MC & 102.34 \\
    LSTM-\cite{fragkiadaki2017motion} & 97.11 \\
    LSTM-CVAE & 96.42 \\
    LSTM-BMS & \textbf{95.63} \\
    \bottomrule
    \end{tabular}
  \caption{Evaluation on the MNIST Sequence dataset.}%
  \label{fig:mnist_eval}
\end{table}

\begin{table*}[!t]
\centering
\resizebox{\textwidth}{!}{
\begin{tabular}{ccccccc}
\toprule
Method & Visual & Error at 1.0(sec) & Error at 2.0(sec) & Error at 3.0(sec) & Error at 4.0(sec) & CLL \\
\midrule
LSTM & x & 1.08 & 2.57 & 4.70 & 7.20 & 134.29 \\
LSTM & RGB & 0.84 & 1.95 & 3.86 & 6.24 & 133.12 \\
DESIRE-SI-IT4 \cite{lee2017desire} & RGB & 1.29 & 2.35 & 3.47 & 5.33 & x \\
LSTM-CVAE & RGB & \textbf{0.71} & 1.86 & 3.39 & 5.06 & 127.51 \\
LSTM-BMS & RGB & 0.80  & \textbf{1.77} & \textbf{3.10} & \textbf{4.62} & \textbf{126.65}   \\
\bottomrule
\end{tabular}
}
\caption{Evaluation on the Stanford Drone dataset. Euclidean distance measured at ($\nicefrac{1}{5}$) resolution.}
%\bernt{can you compare to published results in the literature - would be substantially stronger}}
\label{fig:stanford_drone_eval}
\end{table*}

\section{Experiments}

%\mario{the experimental parts on the different datasets are still missing a clear statement of a) the observations b) the interpretations and how these provide evidence for our story}

We evaluate our models both on synthetic and real data. We choose sequence datasets which display multimodality. In particular, we evaluate on key strokes from MNIST sequence data \cite{mnist_seq} (which can be seen as trajectories in a constrainted space), human trajectories from Stanford Drone data \cite{robicquet2016learning} and radar echo image sequences from HKO \cite{xingjian2015convolutional}. All models were trained using the ADAM optimizer, with a batch size of 32 for trajectory data and 4 for the radar echo data. All experiments were conducted on a single Nvidia M40 GPU with 12GB memory. For models trained using the $\hat{\mathcal{L}}_{\text{CVAE}}$ and $\hat{\mathcal{L}}_{\text{BMS}}$ objectives, we use $T=\left\{10,10,5\right\}$ samples during training on the MNIST Sequence, Stanford Drone, and HKO datasets respectively.

\begin{figure}[h]
\includegraphics[height=4.0cm]{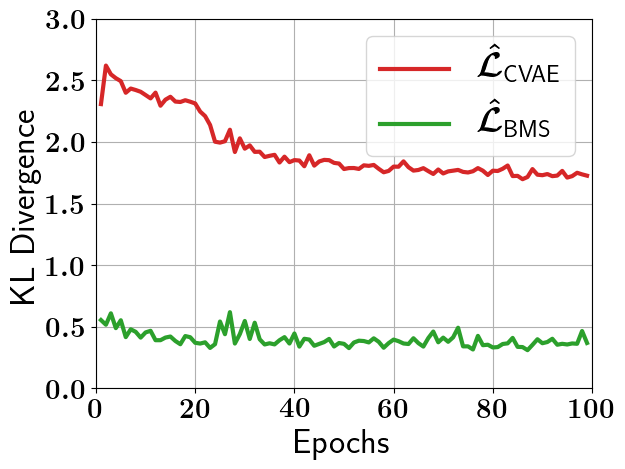}
\caption{KL Divergence during training on the MNIST Sequence dataset.}%
  \label{fig:kl_div}
\end{figure}

\begin{figure*}[t]
  \begin{minipage}{1.0\textwidth}
  \centering
  \begin{tabular}{ c@{\hskip 0.15cm}c@{\hskip 0.15cm}c@{\hskip 0.15cm}c@{\hskip 0.15cm}c }
    
    \includegraphics[height=2.5cm]{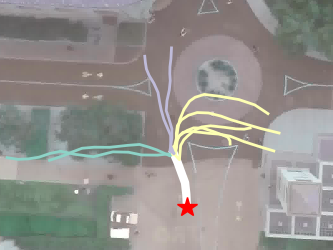} &
    \includegraphics[height=2.5cm]{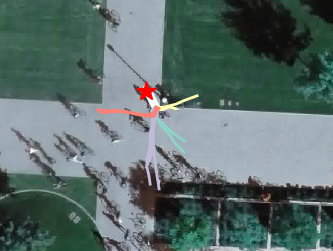} &
    \includegraphics[height=2.5cm]{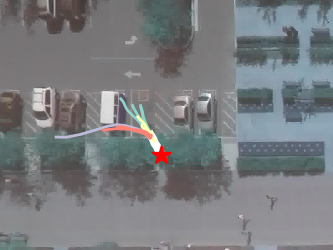} &
    \includegraphics[height=2.5cm]{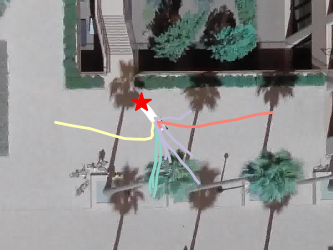} &
    \includegraphics[height=2.5cm]{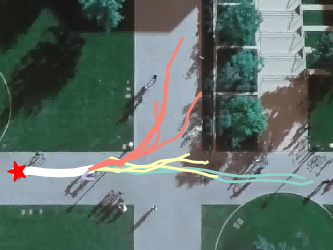} \\

    \includegraphics[height=2.5cm]{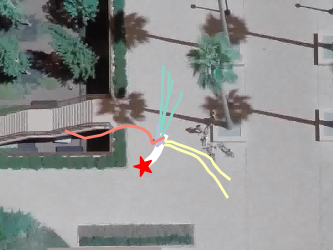} &
    \includegraphics[height=2.5cm]{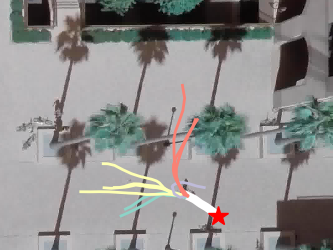} &
    \includegraphics[height=2.5cm]{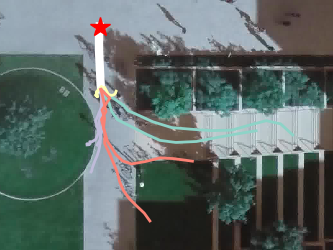} &
    \includegraphics[height=2.5cm]{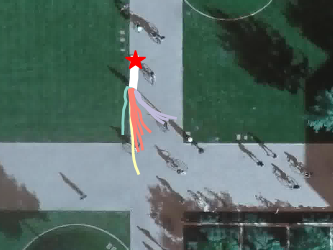} &
    \includegraphics[height=2.5cm]{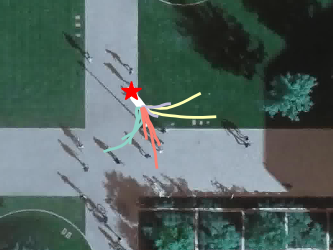} \\
    
    \end{tabular}
  \subcaption{Diverse samples dawn from our LSTM-BMS model trained using the $\hat{\mathcal{L}}_{\text{BMS}}$ objective, color-coded after clustering using k-means with four clusters.}
  \label{fig:stanford_drone_cluster}
  \end{minipage}
  \begin{minipage}{1.0\textwidth}
  \centering
  \begin{tabular}{ c@{\hskip 0.15cm}c@{\hskip 0.15cm}c@{\hskip 0.15cm}c@{\hskip 0.15cm}c }
    
    \includegraphics[height=2.5cm]{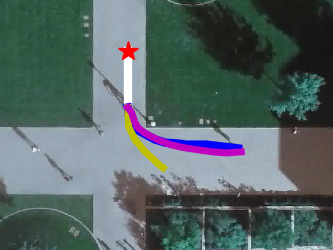} &
    \includegraphics[height=2.5cm]{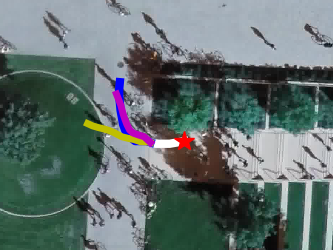} &
    \includegraphics[height=2.5cm]{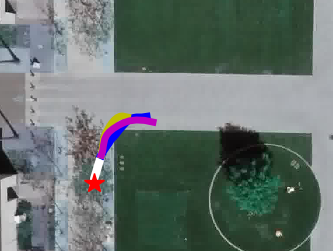} &
    \includegraphics[height=2.5cm]{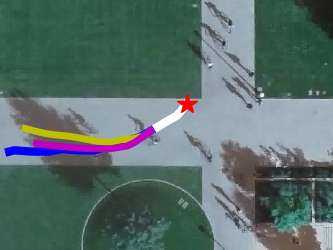} &
    \includegraphics[height=2.5cm]{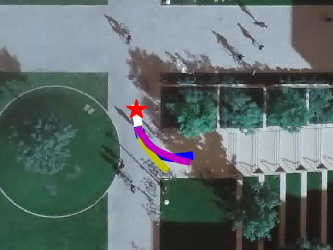} \\

    \includegraphics[height=2.5cm]{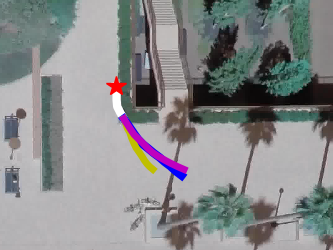} &
    \includegraphics[height=2.5cm]{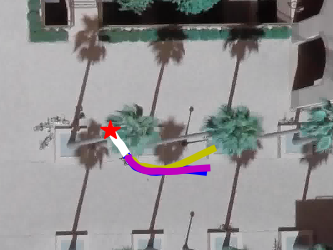} &
    \includegraphics[height=2.5cm]{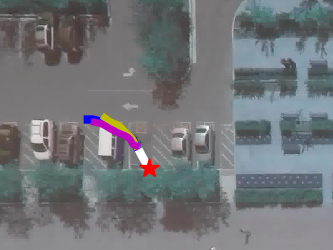} &
    \includegraphics[height=2.5cm]{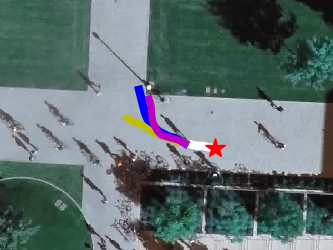} &
    \includegraphics[height=2.5cm]{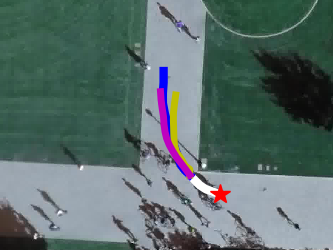} \\
    
    \end{tabular}
  \subcaption{Top 10\% of samples drawn from the LSTM-BMS model (\textcolor{black}{margenta}) and the LSTM-CVAE model (\textcolor{black}{yellow}), with the groundtruth in \textcolor{black}{blue}.}
  \label{fig:stanford_drone_comp}
  \end{minipage}
 \caption{Qualitative evaluation on the Stanford Drone dataset.}
\end{figure*}

\subsection{MNIST Sequence}
The MNIST sequence dataset consists of pen strokes which closely approximates the skeleton of the digits in the MNIST dataset. We focus on the stroke completion task. Given an initial stroke the distribution of possible completions is highly multimodal. The digits 0, 3, 2 and 8, have the same initial stroke with multiple writing styles for each digit. Similarly for the digits 0 and 6, with multiple writing styles for each digit. 

We fix the length of the initial stroke sequence at 10. We use the trajectory prediction model from \autoref{fig:traj_model} and train it using the $\hat{\mathcal{L}}_{\text{BMS}}$ objective (LSTM-BMS). We compare it against the following baselines: \begin{enumerate*} \item A vanilla LSTM encoder-decoder regression model (LSTM) without latent variables; \item The trajectory prediction model from \autoref{fig:traj_model} trained using the $\hat{\mathcal{L}}_{\text{MC}}$ objective (LSTM-MC); \item The trajectory prediction model from \autoref{fig:traj_model} trained using the $\hat{\mathcal{L}}_{\text{CVAE}}$ objective (LSTM-CVAE); \item The trajectory prediction model from \autoref{fig:traj_model} trained using the multi-sample objective of \cite{fragkiadaki2017motion} (LSTM-\cite{fragkiadaki2017motion}) \end{enumerate*}. We use the negative conditional log-likelihood metric (CLL) and report the results in \autoref{fig:mnist_eval}. We use $T=100$ samples to estimate the CLL, following the standard approach of \cite{sohn2015learning}. 

We observe that our LSTM-BMS model achieves the best CLL. This means that our LSTM-BMS model fits the data distribution best. Furthermore, the superior performance compared to the multi-sample of demonstrates the effectiveness of Importance Sampling. Furthermore, we see that the latent variables sampled from our recognition network $q_{\phi}(z \mid x, y)$  during training better matches the true distribution $p(z \mid x)$ used during testing. This can be seen through the KL divergence $\kldiv{q_{\phi}(z \mid x, y)}{p(z \mid x)}$ in \autoref{fig:kl_div} during training of the recognition network trained with the $\hat{\mathcal{L}}_{\text{BMS}}$ objective versus that of the $\hat{\mathcal{L}}_{\text{CVAE}}$ objective. We observe that the KL divergence of the recognition network trained with the $\hat{\mathcal{L}}_{\text{BMS}}$ to be substantially lower, thus, reducing the mismatch in the latent variable $z$ between the training and testing pipelines.

We show qualitative examples of generated samples in \autoref{fig:mnist_ex} from the LSTM-BMS model. We show $T=100$ samples per test example. The initial conditioning stroke is shown in white. The samples drawn are diverse and clearly multimodal. We cluster the generated samples using k-means for better visualization. The number of clusters is set manually to the number of expected digits based on the initial stroke. In particular, our model generates  corresponding to 2, 3, 0 (1st example), 0, 6 (2nd example) and so on.

We compare the accuracy of samples generated by our LSTM-BMS model versus the LSTM-CVAE model in \autoref{fig:mnist_comp}. We display mean of the oracle top 10\% of samples (closest in euclidean distance  w.r.t. groudtruth) generated by both models. Comparing the results we see that, using the $\hat{\mathcal{L}}_{\text{BMS}}$ objective leads to more accurate samples.

\begin{table*}[h]
\centering
\begin{tabular}{ccccccc}
\toprule
Method & Rainfall-MSE & CSI & FAR & POD & Correlation & CLL \\
\midrule
\cite{xingjian2015convolutional} & 1.420 & 0.577 & 0.195 & 0.660 & 0.908 & x \\
Conv-LSTM-CVAE & 1.259  & 0.651 & \textbf{0.155} & 0.701 & 0.910 & 132.78 \\
Conv-LSTM-BMS & \textbf{1.163}  & \textbf{0.670} & 0.163 & \textbf{0.734} & \textbf{0.918} & \textbf{132.52} \\
\bottomrule
\end{tabular}
\caption{Evaluation on HKO radar image sequences.}
\label{fig:hko_eval}
\end{table*}

\begin{figure*}[h]
  
  \centering
  \begin{tabular}{c@{\hskip 0.35cm}c@{\hskip 0.05cm}c@{\hskip 0.05cm}c@{\hskip 0.05cm}c@{\hskip 0.40cm}c@{\hskip 0.05cm}c@{\hskip 0.05cm}c@{\hskip 0.05cm}c}
  \toprule
  	Time-step & Groundtruth & Best & Mean & Variance  & Groundtruth & Best & Mean & Variance \\
  	\midrule
    $t+5$ &
    \includegraphics[height=1.5cm]{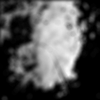} &
    \includegraphics[height=1.5cm]{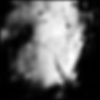} &
    \includegraphics[height=1.5cm]{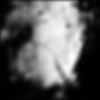} &
    \includegraphics[height=1.5cm,trim={2.5cm 2.5cm 2.5cm 2.5cm},clip]{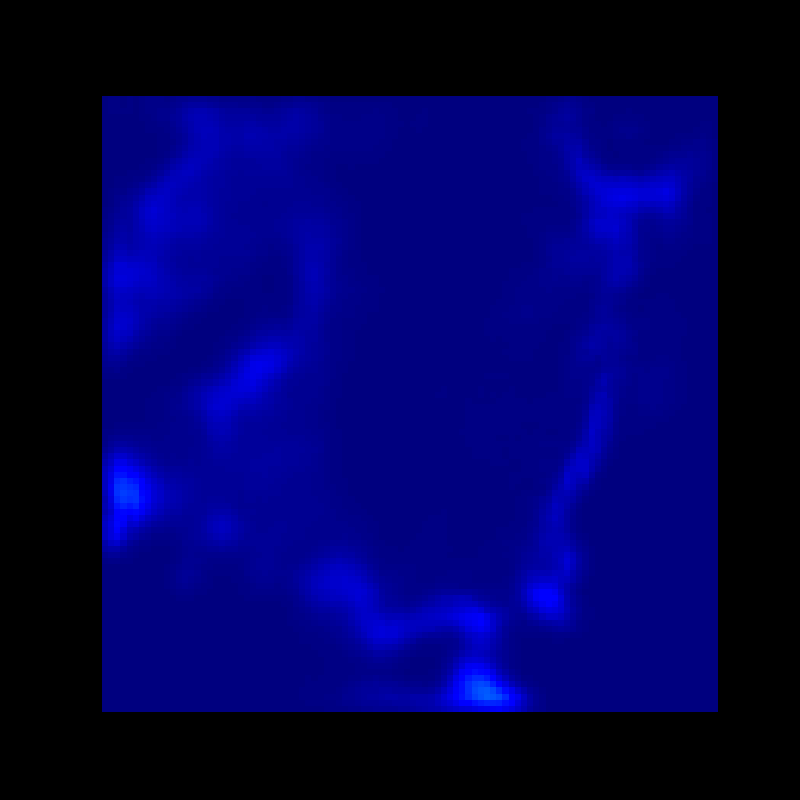} &
    \includegraphics[height=1.5cm]{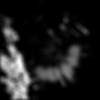} &
    \includegraphics[height=1.5cm]{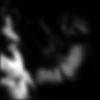} &
    \includegraphics[height=1.5cm]{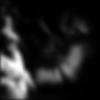} &
    \includegraphics[height=1.5cm,trim={2.5cm 2.5cm 2.5cm 2.5cm},clip]{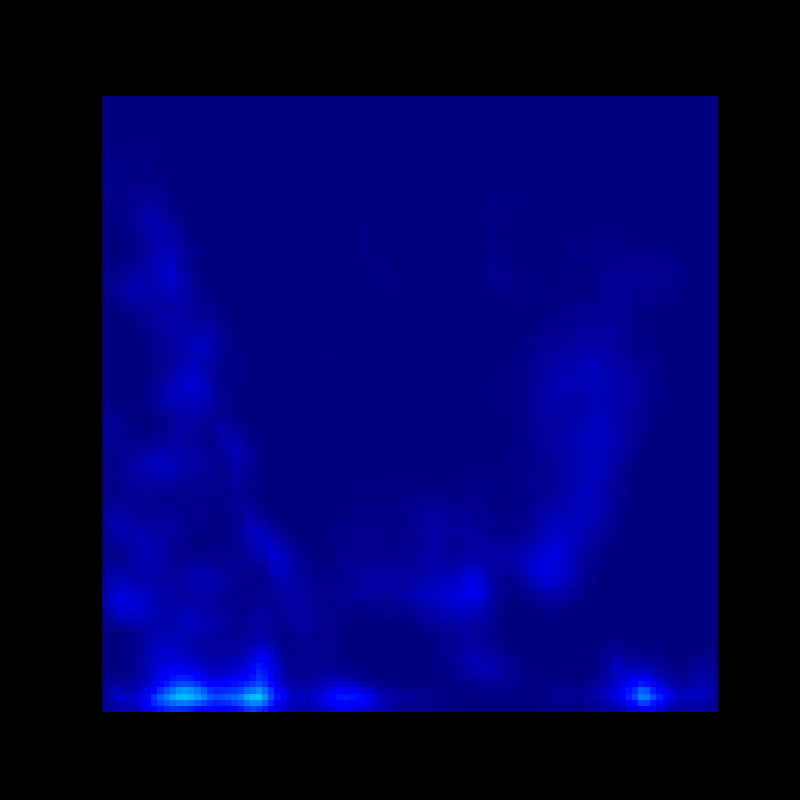} 
    \\
    \midrule
    $t+10$ &
   \includegraphics[height=1.5cm]{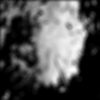} &
    \includegraphics[height=1.5cm]{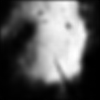} &
    \includegraphics[height=1.5cm]{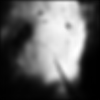} &
    \includegraphics[height=1.5cm,trim={2.5cm 2.5cm 2.5cm 2.5cm},clip]{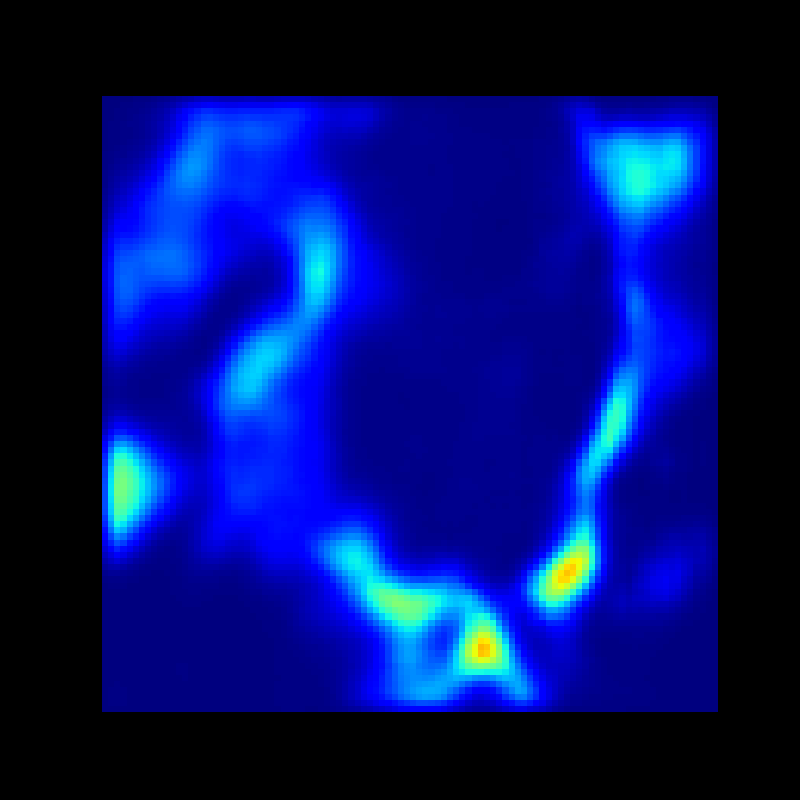} &
    \includegraphics[height=1.5cm]{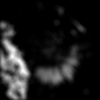} &
    \includegraphics[height=1.5cm]{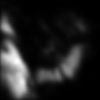} &
    \includegraphics[height=1.5cm]{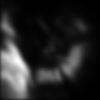} &
    \includegraphics[height=1.5cm,trim={2.5cm 2.5cm 2.5cm 2.5cm},clip]{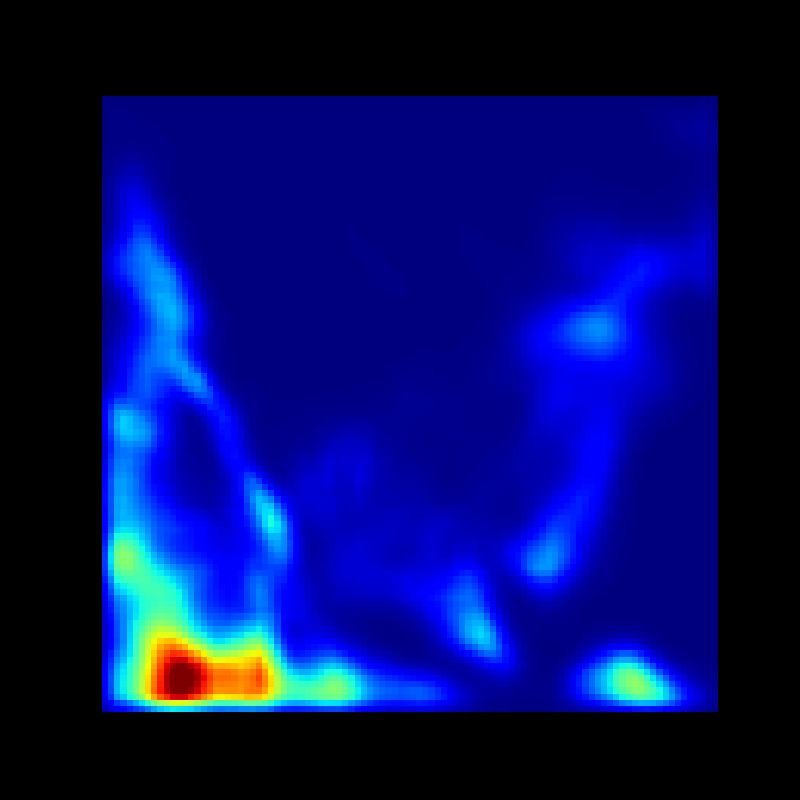} 
    
    \\
    \midrule
    $t+15$ &
    \includegraphics[height=1.5cm]{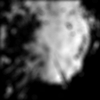} &
    \includegraphics[height=1.5cm]{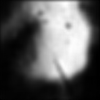} &
    \includegraphics[height=1.5cm]{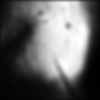} &
    \includegraphics[height=1.5cm,trim={2.5cm 2.5cm 2.5cm 2.5cm},clip]{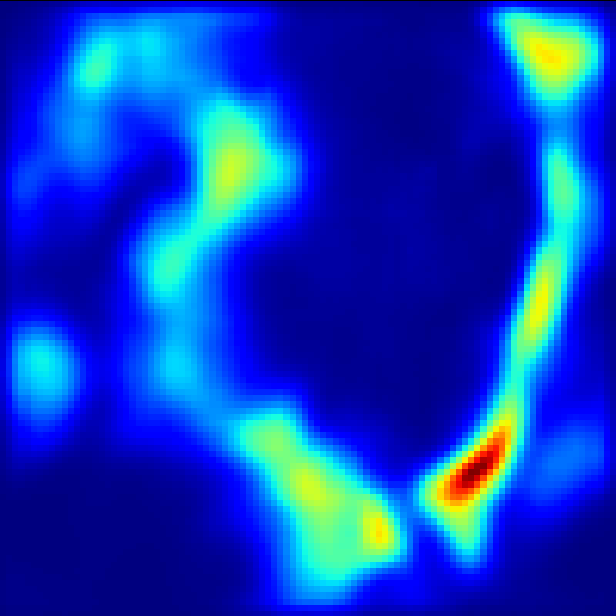} &
    \includegraphics[height=1.5cm]{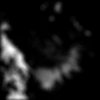} &
    \includegraphics[height=1.5cm]{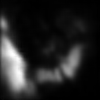} &
    \includegraphics[height=1.5cm]{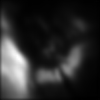} &
    \includegraphics[height=1.5cm,trim={2.5cm 2.5cm 2.5cm 2.5cm},clip]{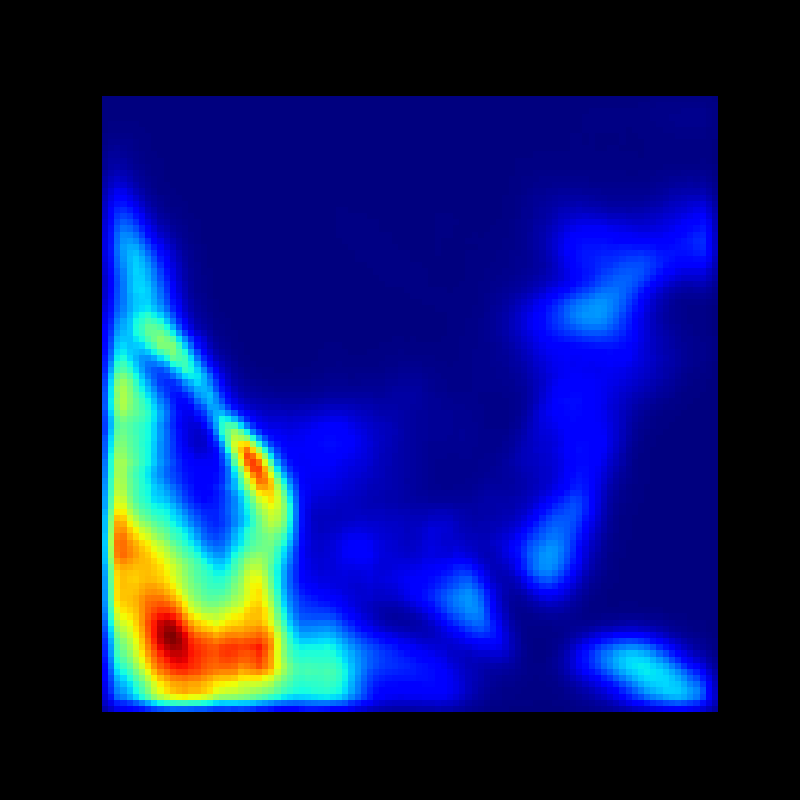} 

	\\
    
    \bottomrule
    
    \end{tabular}
  \caption{Statistics of samples generated by our LSTM-BMS model on the HKO dataset.}
  \label{fig:hko_frames}
\end{figure*}

\subsection{Stanford Drone}
The Stanford Drone dataset consists of overhead videos of traffic scenes. Trajectories of various dynamic agents including Pedestrians and Bikers are annotated. The paths of such agents are determined by various factors including the intention of the agent, paths of other agents and the layout of the scene. Thus, the trajectories are highly multi-modal. As in \cite{robicquet2016learning,lee2017desire}, we predict the trajectories of these agents 4.8 seconds into the future conditioned on the past 2.4 seconds. We use the same evaluation procedure and dataset split as in \cite{lee2017desire}. We encode trajectories as relative displacement from the initial position. The trajectory at each time-step can be seen as the velocity of the agent.

We consider the extension of our trajectory prediction model (\autoref{fig:traj_model}) discussed in \autoref{ssec:models} conditioned on the last visual observation from the overhead camera. We use a 6 layer CNN to extract visual features (see supplementary material). We train this model with the  $\hat{\mathcal{L}}_{\text{BMS}}$ objective and compare it to: \begin{enumerate*} \item A vanilla LSTM encoder-decoder regression model with and without visual observation (LSTM); \item The state of the art DESIRE-SI-IT4 model from \cite{lee2017desire}; \item Our extended trajectory prediction model \autoref{fig:traj_model} trained using the $\hat{\mathcal{L}}_{\text{CVAE}}$ objective (LSTM-CVAE). \end{enumerate*} 

%The first four layers being convolutional and the last two being fully connected. The convolutional layers have 32, 64, 128 and 256 filters respectively. All filters are of size 3$\times$3 and we use max pooling of size 2$\times$2 after every convolutional layer. The last two fully connected layers are of size 1024 and 32 respectively. We use \emph{tanh} non-linearities after every layer. The output of the last fully connected layer and the summary of the LSTM encoder are concatenated.

We report the results in \autoref{fig:stanford_drone_eval}. We report the CLL metric and the euclidean distance in pixels between the true trajectory and the oracle top 10\% of generated samples at 1, 2, 3 and 4 seconds into the future at ($\nicefrac{1}{5}$) resolution (as in \cite{lee2017desire}). Our LSTM-BMS model again performs best both with respect to the euclidean distance and the CLL metric. This again demonstrates that using the $\hat{\mathcal{L}}_{\text{BMS}}$ objective enables us to better fit the groundtruth data distribution and enables the generation of more accurate samples. The performance advantage with respect to DESIRE-SI-IT4 \cite{lee2017desire} is due to  \begin{enumerate*} \item Conditioning the decoder LSTM in \autoref{fig:traj_model} directly on the visual input at higher ($\nicefrac{1}{2}$ versus $\nicefrac{1}{5}$) resolution (as our LSTM-CVAE outperforms DESIRE-SI-IT4 ), \item Our $\hat{\mathcal{L}}_{\text{BMS}}$ objective (as our LSTM-BMS outperforms both DESIRE-SI-IT4 and LSTM-CVAE). \end{enumerate*}

We show qualitative examples of generated samples ($T=10$) in \autoref{fig:stanford_drone_cluster}. We color code the generated samples using k-means with four clusters. The qualitative examples display high plausibility and diversity. They follow the layout of the scene, the location of roads, vegetation, vehicles etc. We qualitatively compare the accuracy of samples generated by our LSTM-BMS model versus the LSTM-CVAE model in \autoref{fig:stanford_drone_comp}. We see that the oracle top 10\% of samples generated using the $\hat{\mathcal{L}}_{\text{BMS}}$ objective are more accurate and thus more representative of the groundtruth. 

%\bernt{do you use the exact same setting as the reference? if not - why? - if yes - say that explicitly}

\subsection{Radar Echo}
The Radar Echo dataset \cite{xingjian2015convolutional} consists of weather radar intensity images from 97 rainy days over Hong Kong from 2011 to 2013. The weather evolves due to varity of factors, which are difficult to identify using only the radar images, with varied and multimodal futures. Each sequences consists of 20 frames each of resolution 100$\times$100, recorded at intervals of 6 minutes. We use the same dataset split as \cite{xingjian2015convolutional} and predict the next 15 images given the previous 5 images. 

We compare our image sequence prediction model in \autoref{fig:imseq_model} trained with the $\hat{\mathcal{L}}_{\text{BMS}}$ (Conv-LSTM-BMS) objective to one trained with the $\hat{\mathcal{L}}_{\text{CVAE}}$ (Conv-LSTM-CVAE) objective. We additionally compare it to the Conv-LSTM model of \cite{xingjian2015convolutional}. In addition to the CLL metric (calculated per image sequence), we use the following precipitation nowcasting metrics from \cite{xingjian2015convolutional}, \begin{enumerate*} \item Rainfall mean squared error (Rainfall-MSE), \item Critical success index (CSI), \item False alarm rate (FAR), \item Probability of detection (POD), and \item Correlation \end{enumerate*}. For fair comparison we estimate these metrics using $T=1$ random samples from the Conv-LSTM-CVAE and Conv-LSTM-BMS models.

We report the results in \autoref{fig:hko_eval}. Both the Conv-LSTM-CVAE and Conv-LSTM-BMS models perform better compared to \cite{xingjian2015convolutional}. This is due to use of embedding layers for feature extraction and the use of 2$\times$2 max pooling in between two Conv-LSTM layers for feature aggregation (compared no embedding layers or pooling in \cite{xingjian2015convolutional}).
Furthermore, the superior CLL of the Conv-LSTM-BMS model demonstrates it's ability to fit the data distribution better. We show qualitative examples in \autoref{fig:hko_frames} at $t+5$, $t+10$ and $t+15$. We generate $T=50$ samples and show the sample closest to the groundtruth (Best), the mean of all the samples and the per-pixel variance in the samples. The qualitative examples demonstrate that our model produces highly accurate and diverse samples.

\section{Conclusion}
We have presented a novel ``best of many'' sample objective for Gaussian latent variable models and show its advantages for learning conditional models on multi-modal distributions. Our analysis shows indeed the learnt latent representation is better matched between training and test time -- which in turn leads to more accurate samples. We show the benefits of our model on trajectory as well as image sequence prediction using three diverse datasets: MNIST strokes, Stanford drone and HKO weather. Our proposed appraoch consistently outperforms baselines and state of the art in all these scenarios.

\myparagraph{Acknowledgments} We would like to thank Francesco Croce for his comments and feedback.

% -----------------------------------------------------------------------------------

{\small
\bibliographystyle{ieee}
\bibliography{egbib}
}

\newpage

\begin{appendix}
\addappheadtotoc

\title{
   \begin{center}
      \Large\textbf{Appendix}\\
      %\large\textit{A. Thor}
   \end{center}
}

\maketitle

\section{Additional Details of our ``Best of Many'' Sample Objective}
Here we provide additional details of our ``Best of Many'' samples objective and include additional qualitative results.
We begin with the formal statement of the First Mean Value Theorem of Integration \cite{comenetz2002calculus}. The First Mean Value Theorem of Integration states that, if $f_{1}: [a,b] \to \mathbb{R}$ is continuous and $f_{2}$ is an integrable function that does not change sign on $[a,b]$, then $\exists z^{\prime} \in (a,b)$ such that,
\begin{align}\tag{S1}
\int_{a}^{b} \, f_{1}(z) \, f_{2}(z) \, dz = f_{1}(z^{\prime}) \,\,\,  \int_{a}^{b} f_{2}(z) \, dz
\end{align}

The data log-likelihood Equation (3) in the main paper, estimated using importance sampling using a recognition network $q_{\phi}$ is given by,
\begin{align}\tag{S2}\label{eq9}
\begin{split}
&\log(p_{\theta}(y \mid x)) = \\
 & \log\Big( \int p_{\theta}(y | z, x) \, \frac{p(z | x)}{q_{\phi}(z | x, y)} \, q_{\phi}(z | x, y) \, dz \, \Big).
\end{split}
\end{align}
We apply the First Mean Value Theorem of Integration to derive Equation (4) in the main paper, which is,
\begin{align}\tag{S3}\label{eq10}
\begin{split}
\log(p_{\theta}(y | x)) &= \log\Big( \int_{a}^{b} p_{\theta}(y | z, x) \, q_{\phi}(z | x, y) \, dz \, \Big) \\
&+ \log\Big( \frac{p(z^{\prime} | x)}{q_{\phi}(z^{\prime} | x, y)} \Big), \,\, z^{\prime} \in (a,b).
\end{split}
\end{align} 
To do this, we set $f_{1}(z) = \nicefrac{p(z | x)}{q_{\phi}(z | x, y)}$ and $f_{2}(z) = p_{\theta}(y | z, x) \times q_{\phi}(z | x, y)$ (from the data log-likelihood in (\ref{eq9})). The integral in (\ref{eq9}) can be very well approximated on a large enough bounded interval $[a,b]$. This leads to,
\begin{align}\tag{S4}\label{eq3}
\begin{split}
 &\Big( \int_{a}^{b} p_{\theta}(y | z, x) \, \frac{p(z | x)}{q_{\phi}(z | x, y)} \, q_{\phi}(z | x, y) \, dz \, \Big)\\
 &= \frac{p(z^{\prime} | x)}{q_{\phi}(z^{\prime} | x, y)} \,\, \Big( \int_{a}^{b} p_{\theta}(y | z, x) \, q_{\phi}(z | x, y) \, dz \, \Big). 
\end{split}
\end{align}
Taking $\log$ on both sizes of (\ref{eq3}) leads to (\ref{eq10}). We can further lower bound (\ref{eq10}), leading to Equation (5) in the main paper, which is,
\begin{align}\tag{S5}\label{eq11}
\begin{split}
\log(p_{\theta}(y | x)) &\geq \log\Big( \int_{a}^{b} p_{\theta}(y | z, x) \, q_{\phi}(z | x, y) \, dz \, \Big) \\
&+ \min_{z^{\prime} \in (a,b)}\Big( \log\Big( \frac{p(z^{\prime} | x)}{q_{\phi}(z^{\prime} | x, y)} \Big)\Big)
\end{split}
\end{align}
However, as mentioned in the main paper, the minimum in (\ref{eq11}) is difficult to estimate. Therefore, we use the following approximation. From (\ref{eq10}), we know that $\exists z^{\prime} \,\, \in (a,b)$ which lower bounds the data log-likelihood. To maximize this data log-likelihood, we would like to maximize $\log(f_{1}(z^{\prime}))$. However, as we do not know $z^{\prime}$, we instead choose to maximize it for a set of $N$ points in $(a,b)$,
\begin{align}\tag{S6}
\begin{split}
&\log\Big( \int_{a}^{b} p_{\theta}(y | z, x) \, q_{\phi}(z | x, y) \, dz \, \Big) \\
&+ \log\Big( \frac{p(z_{1}^{\prime} | x)}{q_{\phi}(z_{1}^{\prime} | x, y)} \Big) + .. + \log\Big( \frac{p(z_{N}^{\prime} | x)}{q_{\phi}(z_{N}^{\prime} | x, y)} \Big).
\end{split}
\end{align}
As values of both $p$ and $q_{\phi}$ are bounded above by 1, the value of the function $f_{2}(z_{i}^{\prime}) = \nicefrac{p(z_{i}^{\prime} | x)}{q_{\phi}(z_{i}^{\prime} | x, y)}$ is likely to be low when is $p$ low and $q_{\phi}$ is high. Therefore, to give more importance to such points $z_{i}^{\prime}$, we weight each point by $q_{\phi}(z_{i}^{\prime} | x, y)$,
\begin{align}\tag{S7}\label{eq4}
\begin{split}
&\log\Big( \int_{a}^{b} p_{\theta}(y | z, x) \, q_{\phi}(z | x, y) \, dz \, \Big) \\
&+ q_{\phi}(z_{1}^{\prime} | x, y) \times \log\Big( \frac{p(z_{1}^{\prime} | x)}{q_{\phi}(z_{1}^{\prime} | x, y)} \Big) \\
&+ \ldots + q_{\phi}(z_{N}^{\prime} | x, y) \times \log\Big( \frac{p(z_{N}^{\prime} | x)}{q_{\phi}(z_{N}^{\prime} | x, y)} \Big).
\end{split}
\end{align}
Flipping the sign before the terms in the second part of (\ref{eq4}),
\begin{align}\tag{S8}\label{eq5}
\begin{split}
&\log\Big( \int_{a}^{b} p_{\theta}(y | z, x) \, q_{\phi}(z | x, y) \, dz \, \Big) \\
&- q_{\phi}(z_{1}^{\prime} | x, y) \times \log\Big( \frac{q_{\phi}(z_{1}^{\prime} | x, y)}{p(z_{1}^{\prime} | x)} \Big) \\
&- \ldots - q_{\phi}(z_{N}^{\prime} | x, y) \times \log\Big( \frac{q_{\phi}(z_{N}^{\prime} | x, y)}{p(z_{N}^{\prime} | x)} \Big).
\end{split}
\end{align}
If we choose a sufficiently large set of points $z_{i}^{\prime} \in (a,b)$, we can collect the terms in the second part of (\ref{eq5}) and replace them with a single integral,
\begin{align}\tag{S9}\label{eq6}
\begin{split}
&\log\Big( \int_{a}^{b} p_{\theta}(y | z, x) \, q_{\phi}(z | x, y) \, dz \, \Big) \\
&- \int_{a}^{b} \, q_{\phi}(z | x, y) \times \log\Big( \frac{q_{\phi}(z | x, y)}{p(z_ | x)} \Big) \,\, dz. 
\end{split}
\end{align}
The second integral in (\ref{eq6}) is the KL divergence between the two distributions $q_{\phi}(z | x, y)$ and $p(z_ | x)$,
\begin{align}\tag{S10}\label{eq7}
\begin{split}
&\log\Big( \int_{a}^{b} p_{\theta}(y | z, x) \, q_{\phi}(z | x, y) \, dz \, \Big) \\
&- \kldiv{q_{\phi}(z | x, y)}{p(z | x)}.
\end{split}
\end{align}
We can estimate the data log-likelihood term in (\ref{eq7}) using Monte-Carlo integration. This leads to the ``Many Sample'' objective from the main paper,
\begin{align}\tag{S11}\label{eq8}
\begin{split}
&\hat{\mathcal{L}}_{\text{MS}} = \log\Big( \frac{1}{T} \sum_{i=1}^{T} p_{\theta}(y | \hat{z}_{i}, x) \, \Big) \\
&- \kldiv{q_{\phi}(z | x, y)}{p(z | x)}, \,\,  \hat{z}_{i} \sim q_{\phi}(z | x, y).
\end{split}
\end{align}
As mentioned in the main paper, we use the re-parameterization trick \cite{kingma2013auto} to sample from our recognition network $q_{\phi}$. Therefore, the recognition network predicts the mean and variance $\mathcal{N}(\mu,\sigma)$ of the Gaussian distribution $q_{\phi}$ from which the latent variable $z$ is sampled. Thus, we can directly use the predicted $\mu,\sigma$ to estimate the KL divergence as in \cite{kingma2013auto}.

Approximating the data log-likelihood term in the first part of (\ref{eq8}) as shown in the main paper, leads to our ``Best of Many'' sample objective.

\section{Additional Details of our Models}
Here, we include details of each layer of our models.
\subsection{Model for Structured Trajectory Prediction}\label{ssec:mtraj}  
We provide the details of our structured trajectory prediction model in \autoref{tab:traj_details1}. Followed by the details of the recognition network ($q_{\phi}$) in \autoref{tab:traj_details2}. We refer to fully connected layers as Dense and Size refers to the number of neurons in the layer.
\begin{table}[h]
\centering
\resizebox{\linewidth}{!}{
\begin{tabular}{cccccc}
\toprule
Layer & Type & Size & Activation & Input & Output \\
\midrule
$\text{In}_1$ & Input & & & $x$ & $\text{EMB}_{1}$ \\
\midrule
$\text{EMB}_{1}$ & Dense & 32 & \emph{ReLU} & $\text{In}_1$ & $\text{LSTM}_{enc}$ \\
$\text{LSTM}_{enc}$ & LSTM & 48 & \emph{tanh} & $\text{EMB}_{1}$ & $\text{EMB}_{2}$ \\
\midrule
$\text{EMB}_{2}$ & Dense & 64 & \emph{ReLU} & $\left\{\text{LSTM}_{enc},q_{\phi}\right\}$ & $\text{LSTM}_{dec}$ \\
$\text{LSTM}_{dec}$ & LSTM & 48 & \emph{tanh} & $\text{EMB}_{2}$ & $\text{Out}_{1}$ \\
$\text{Out}_{1}$ & Dense & 2 & & $\text{LSTM}_{dec}$ & $\hat{y}$ \\
\bottomrule
\end{tabular}
}
\caption{Details our model for Structured Trajectory Prediction. The details of the recognition network $q_{\phi}$ used during training follows in \autoref{tab:traj_details2}.}
\label{tab:traj_details1}
\end{table}

\begin{table}[h]
\centering
\resizebox{\linewidth}{!}{
\begin{tabular}{cccccc}
\toprule
Layer & Type & Size & Activation & Input & Output \\
\midrule
$\text{In}_2$ & Input & & & $y$ & $\text{EMB}_{3}$ \\
\midrule
$\text{EMB}_{3}$ & Dense & 64 & \emph{ReLU} & $\text{In}_2$ & $\text{LSTM}_{rec}$ \\
$\text{LSTM}_{rec}$ & LSTM & 128 & \emph{tanh} & $\text{EMB}_{3}$ & $\left\{\text{D}_{1},\text{D}_{2}\right\}$ \\
$\text{D}_{1}$ & Dense & 64 & & $\text{LSTM}_{rec}$ & $\mu$ \\
$\text{D}_{2}$ & Dense & 64 & & $\text{LSTM}_{rec}$ & $\sigma$ \\
\bottomrule
\end{tabular}
}
\caption{Details of the recognition network used during training of our model for Structured Trajectory Prediction.}
\label{tab:traj_details2}
\end{table}

\FloatBarrier
\subsection{Extension with Visual Input}
This model is similar to the model for Structured Trajectory Prediction, expect that the $\text{LSTM}_{dec}$ is additionally conditioned on the output of an CNN encoder. The details are in \autoref{tab:cnne_details1} and \autoref{tab:vis_details1}. We use the same recognition network as described previously in \autoref{ssec:mtraj}.

\begin{table}[h]
\centering
\resizebox{\linewidth}{!}{
\begin{tabular}{ccccccc}
\toprule
Layer & Type & Filters & Size & Activation & Input & Output \\
\midrule
$\text{In}_2$ & Input & & & & & $\text{C}_1$ \\
\midrule
$\text{C}_1$ & Conv & 32 & 3$\times$3 & \emph{tanh} & $\text{In}_2$ & $\text{P}_1$ \\
$\text{P}_1$ & MaxPool & & 2$\times$2 & & $\text{C}_1$ & $\text{C}_2$ \\
$\text{C}_2$ & Conv & 64 & 3$\times$3 & \emph{tanh} & $\text{P}_1$ & $\text{P}_2$ \\
$\text{P}_2$ & MaxPool & & 2$\times$2 & & $\text{C}_2$ & $\text{C}_3$ \\
$\text{C}_3$ & Conv & 128 & 3$\times$3 & \emph{tanh} & $\text{P}_2$ & $\text{P}_3$ \\
$\text{P}_3$ & MaxPool & & 2$\times$2 & & $\text{C}_3$ & $\text{C}_4$ \\
$\text{C}_4$ & Conv & 256 & 3$\times$3 & \emph{tanh} & $\text{P}_3$ & $\text{P}_4$ \\
$\text{P}_4$ & MaxPool & & 2$\times$2 & & $\text{C}_4$ & $\text{FC}_1$ \\
\midrule
$\text{FC}_1$ & Dense & 1024 &  & \emph{tanh} & $\text{P}_4$ & $\text{FC}_2$ \\
$\text{FC}_2$ & Dense & 32 & & \emph{tanh} & $\text{FC}_1$ & $\text{EMB}_{2}$ \\
\bottomrule
\end{tabular}
}
\caption{Details of the CNN encoder used with the extended Structured Trajectory Prediction model with Visual Input. Conv stands for 2D convolution, MaxPool stands for 2D max pooling and UpSample stands for 2D upsampling operations.}
\label{tab:cnne_details1}
\end{table}

\begin{table}[h]
\centering
\resizebox{\linewidth}{!}{
\begin{tabular}{cccccc}
\toprule
Layer & Type & Size & Activation & Input & Output \\
\midrule
$\text{In}_1$ & Input & & & $x$ & $\text{EMB}_{1}$ \\
\midrule
$\text{EMB}_{1}$ & Dense & 32 & \emph{ReLU} & $\text{In}_1$ & $\text{LSTM}_{enc}$ \\
$\text{LSTM}_{enc}$ & LSTM & 48 & \emph{tanh} & $\text{EMB}_{1}$ & $\text{EMB}_{2}$ \\
\midrule
$\text{EMB}_{2}$ & Dense & 64 & \emph{ReLU} & $\left\{\text{LSTM}_{enc},\text{FC}_{2}\right\}$ & $\text{EMB}_{3}$ \\
$\text{EMB}_{3}$ & Dense & 64 & \emph{ReLU} & $\left\{\text{EMB}_{2},q_{\phi}\right\}$ & $\text{LSTM}_{dec}$ \\
$\text{LSTM}_{dec}$ & LSTM & 64 & \emph{tanh} & $\text{EMB}_{3}$ & $\text{Out}_{1}$ \\
$\text{Out}_{1}$ & Dense & 2 & & $\text{LSTM}_{dec}$ & $\hat{y}$ \\
\bottomrule
\end{tabular}
}
\caption{Details our model for extended Structured Trajectory Prediction model with Visual Input. The details of the recognition network $q_{\phi}$ used during training follows in \autoref{tab:cnne_details2}.}
\label{tab:vis_details1}
\end{table}

\begin{table}[h]
\centering
\resizebox{\linewidth}{!}{
\begin{tabular}{cccccc}
\toprule
Layer & Type & Size & Activation & Input & Output \\
\midrule
$\text{In}_3$ & Input & & & $y$ & $\text{EMB}_{4}$ \\
\midrule
$\text{EMB}_{4}$ & Dense & 64 & \emph{ReLU} & $\text{In}_3$ & $\text{LSTM}_{rec}$ \\
$\text{LSTM}_{rec}$ & LSTM & 128 & \emph{tanh} & $\text{EMB}_{3}$ & $\left\{\text{D}_{1},\text{D}_{2}\right\}$ \\
$\text{D}_{1}$ & Dense & 64 & & $\text{LSTM}_{rec}$ & $\mu$ \\
$\text{D}_{2}$ & Dense & 64 & & $\text{LSTM}_{rec}$ & $\sigma$ \\
\bottomrule
\end{tabular}
}
\caption{Details of the recognition network used during training of our extended Structured Trajectory Prediction model with Visual Input.}
\label{tab:cnne_details2}
\end{table}

\FloatBarrier
\subsection{Model for Structured Image Sequence Prediction}
We provide the details of our structured image sequence prediction model in \autoref{tab:imseq_details1}. Followed by the details of the recognition network ($q_{\phi}$) in \autoref{tab:imseq_details2}. In contrast to the model for structured trajectory prediction, we use Convolutional LSTMs and Convolutional Embedding layers. 

\begin{table}[h]
\centering
\resizebox{\linewidth}{!}{
\begin{tabular}{cccccc}
\toprule
Layer & Type & Filters & Size & Input & Output \\
\midrule
$\text{In}_1$ & Input & & & $x$ & $\text{CEMB}_{1}$ \\
\midrule
$\text{CEMB}_{1}$ & Conv & 32 & 3$\times$3 & $\text{In}_1$ & $\text{P}_{1}$ \\
$\text{P}_{1}$ & MaxPool & & 2$\times$2  & $\text{CEMB}_{1}$ & $\text{CLSTM}_{enc1}$ \\
$\text{CLSTM}_{enc1}$ & CLSTM & 32 & 3$\times$3 & $\text{P}_{1}$ & $\text{P}_{2}$ \\
$\text{P}_{2}$ & MaxPool & & 2$\times$2  & $\text{CLSTM}_{enc1}$ & $\text{CLSTM}_{enc2}$ \\
$\text{CLSTM}_{enc2}$ & CLSTM & 64 & 3$\times$3 & $\text{P}_{2}$ & $\text{CEMB}_{2}$ \\
\midrule
$\text{CEMB}_{2}$ & Conv & 32 & 3$\times$3 & $\left\{\text{CLSTM}_{enc2},q_{\phi}\right\}$ & $\text{CLSTM}_{dec1}$ \\
$\text{CLSTM}_{dec1}$ & CLSTM & 64 & 3$\times$3 & $\text{CEMB}_{2}$ & $\text{U}_{1}$ \\
$\text{U}_{1}$ & UpSample & & 2$\times$2  & $\text{CLSTM}_{dec1}$ & $\text{CLSTM}_{dec2}$ \\
$\text{CLSTM}_{dec2}$ & CLSTM & 64 & 3$\times$3 & $\text{U}_{1}$ & $\text{U}_{2}$ \\
$\text{U}_{2}$ & UpSample & & 2$\times$2  & $\text{CLSTM}_{dec2}$ & $\text{Out}_{1}$ \\
$\text{Out}_{1}$ & Conv & 32 & 3$\times$3 & $\text{U}_{2}$ & $\text{Out}_{2}$ \\
$\text{Out}_{2}$ & Conv & 1 & 3$\times$3 & $\text{Out}_{1}$ & $\hat{y}$ \\
\bottomrule
\end{tabular}
}
\caption{Details our model for Structured Image Sequence Prediction. CLSTM stands for 2D Convolutional LSTM, Conv stands for 2D convolution, MaxPool stands for 2D max pooling and UpSample stands for 2D upsampling operations. The details of the recognition network $q_{\phi}$ used during training follows in \autoref{tab:imseq_details2}.}
\label{tab:imseq_details1}
\end{table}

\begin{table}[h]
\centering
\resizebox{\linewidth}{!}{
\begin{tabular}{cccccc}
\toprule
Layer & Type & Filters & Size & Input & Output \\
\midrule
$\text{In}_2$ & Input & & & $y$ & $\text{CEMB}_{3}$ \\
\midrule
$\text{CEMB}_{3}$ & Conv & 32 & 3$\times$3 & $\text{In}_2$ & $\text{P}_{3}$ \\
$\text{P}_{3}$ & MaxPool & & 2$\times$2  & $\text{CEMB}_{3}$ & $\text{CLSTM}_{rec1}$ \\
$\text{CLSTM}_{rec1}$ & CLSTM & 32 & 3$\times$3 & $\text{P}_{3}$ & $\text{P}_{4}$ \\
$\text{P}_{4}$ & MaxPool & & 2$\times$2  & $\text{CLSTM}_{rec1}$ & $\text{CLSTM}_{rec2}$ \\
$\text{CLSTM}_{rec2}$ & CLSTM & 64 & 3$\times$3 & $\text{P}_{4}$ & $\left\{\text{C}_{1},\text{C}_{2}\right\}$ \\
$\text{C}_{1}$ & Conv & 64 & 3$\times$3 & $\text{CLSTM}_{rec2}$ & $\mu$ \\
$\text{C}_{2}$ & Conv & 64 & 3$\times$3 & $\text{CLSTM}_{rec2}$ & $\sigma$ \\
\bottomrule
\end{tabular}
}
\caption{Details of the recognition network used during training of our model for Structured Image Sequence Prediction. CLSTM stands for 2D Convolutional LSTM, Conv stands for 2D convolution, MaxPool stands for 2D max pooling and UpSample stands for 2D upsampling operations.}
\label{tab:imseq_details2}
\end{table}

\FloatBarrier
\section{Additional Results}
We show additional qualitative results on the HKO dataset in \autoref{fig:hko_frames} at $t+5$, $t+10$ and $t+15$. We generate $T=50$ samples and show the sample closest to the groundtruth (Best), the mean of all the samples and the per-pixel variance in the samples. As in the main paper, the qualitative examples demonstrate that our model produces samples which are close to the groundtruth (comparing the Best sample and the groundtruth) and diverse samples (comparing the difference between the mean of the samples and the Best sample). 

\begin{figure*}[h]
  
  \centering
  \begin{tabular}{c@{\hskip 0.35cm}c@{\hskip 0.05cm}c@{\hskip 0.05cm}c@{\hskip 0.05cm}c@{\hskip 0.40cm}c@{\hskip 0.05cm}c@{\hskip 0.05cm}c@{\hskip 0.05cm}c}
  \toprule
  	Time-step & Groundtruth & Best & Mean & Variance  & Groundtruth & Best & Mean & Variance \\
  	\midrule
    $t+5$ &
    \includegraphics[height=1.5cm]{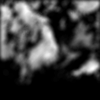} &
    \includegraphics[height=1.5cm]{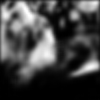} &
    \includegraphics[height=1.5cm]{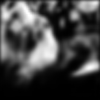} &
    \includegraphics[height=1.5cm,trim={2.5cm 2.5cm 2.5cm 2.5cm},clip]{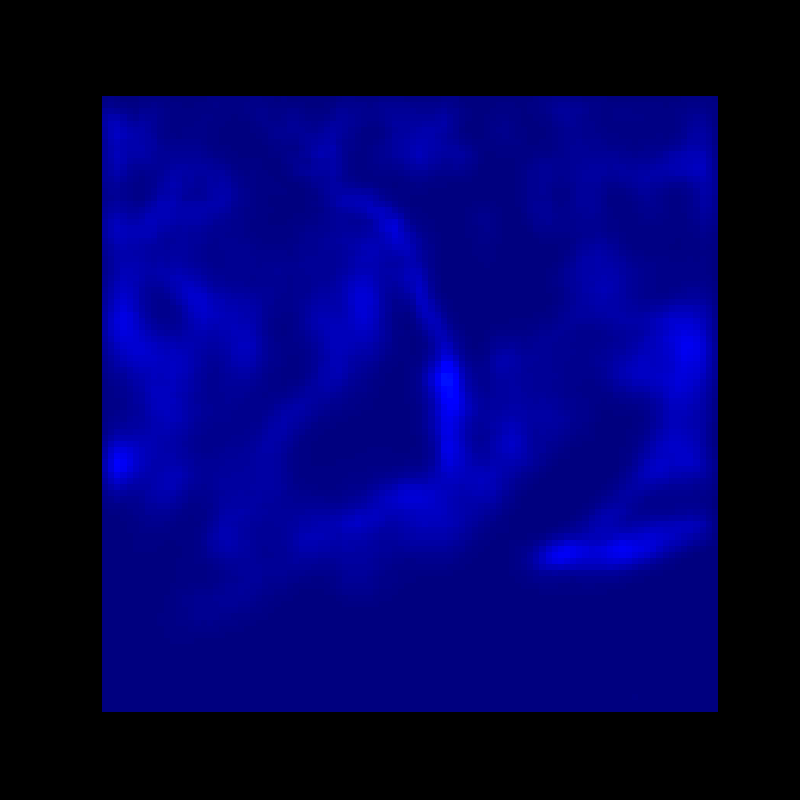} &
    \includegraphics[height=1.5cm]{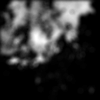} &
    \includegraphics[height=1.5cm]{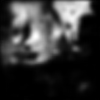} &
    \includegraphics[height=1.5cm]{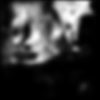} &
    \includegraphics[height=1.5cm,trim={2.5cm 2.5cm 2.5cm 2.5cm},clip]{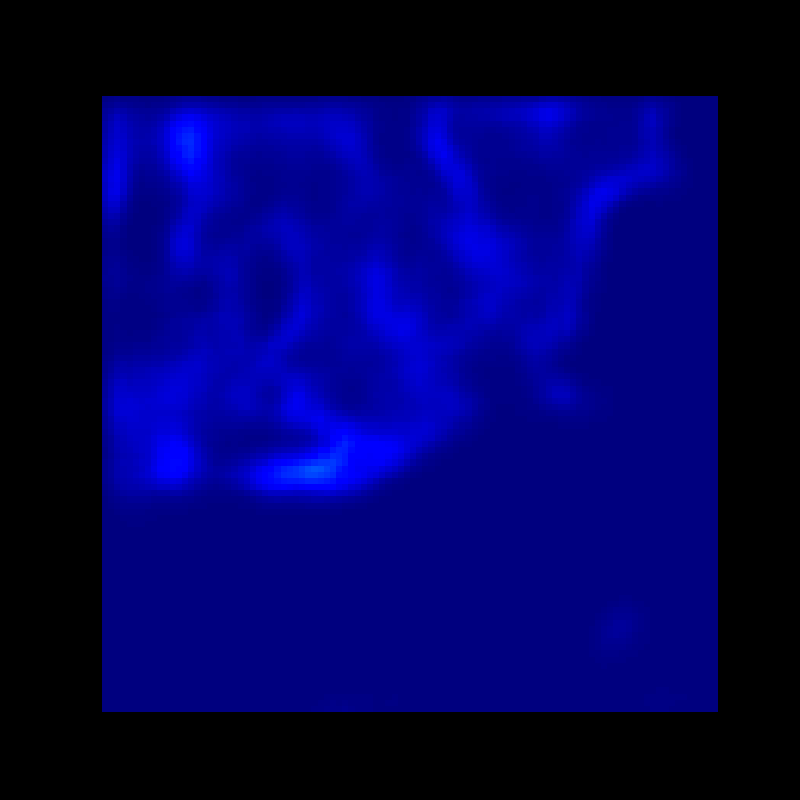} 
    \\
    \midrule
    $t+10$ &
   \includegraphics[height=1.5cm]{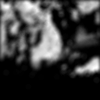} &
    \includegraphics[height=1.5cm]{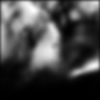} &
    \includegraphics[height=1.5cm]{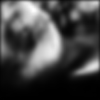} &
    \includegraphics[height=1.5cm,trim={2.5cm 2.5cm 2.5cm 2.5cm},clip]{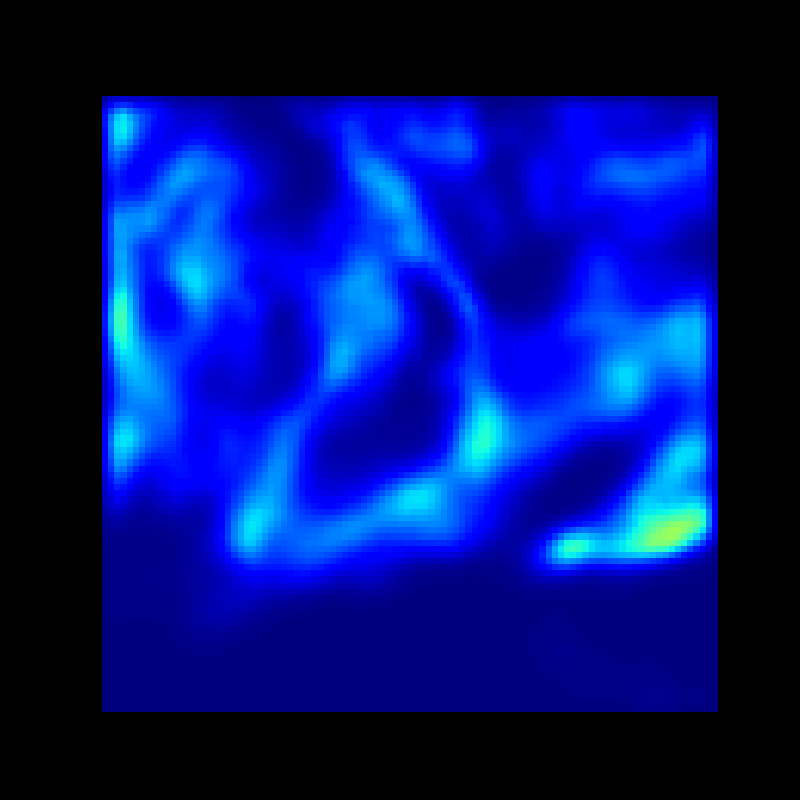} &
    \includegraphics[height=1.5cm]{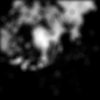} &
    \includegraphics[height=1.5cm]{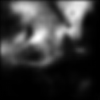} &
    \includegraphics[height=1.5cm]{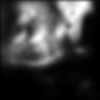} &
    \includegraphics[height=1.5cm,trim={2.5cm 2.5cm 2.5cm 2.5cm},clip]{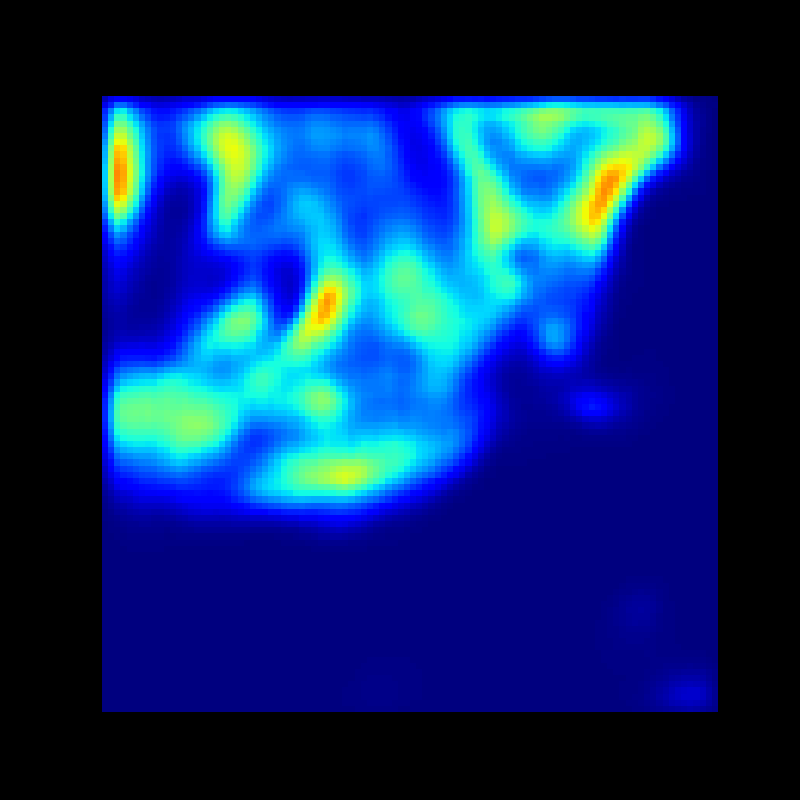} 
    
    \\
    \midrule
    $t+15$ &
    \includegraphics[height=1.5cm]{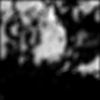} &
    \includegraphics[height=1.5cm]{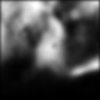} &
    \includegraphics[height=1.5cm]{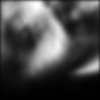} &
    \includegraphics[height=1.5cm,trim={2.5cm 2.5cm 2.5cm 2.5cm},clip]{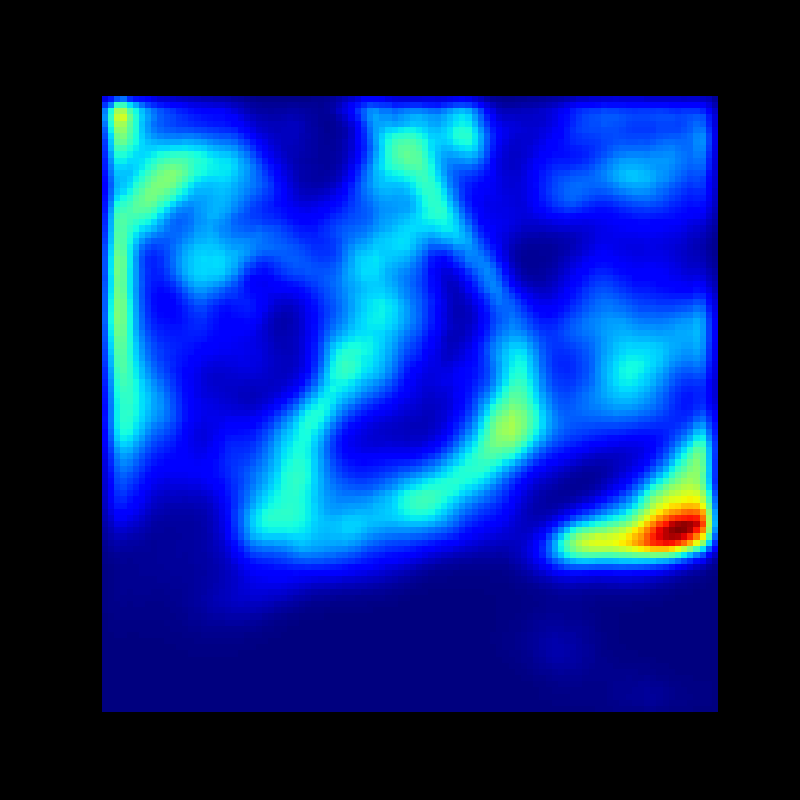} &
    \includegraphics[height=1.5cm]{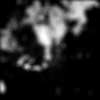} &
    \includegraphics[height=1.5cm]{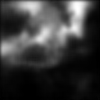} &
    \includegraphics[height=1.5cm]{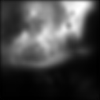} &
    \includegraphics[height=1.5cm,trim={2.5cm 2.5cm 2.5cm 2.5cm},clip]{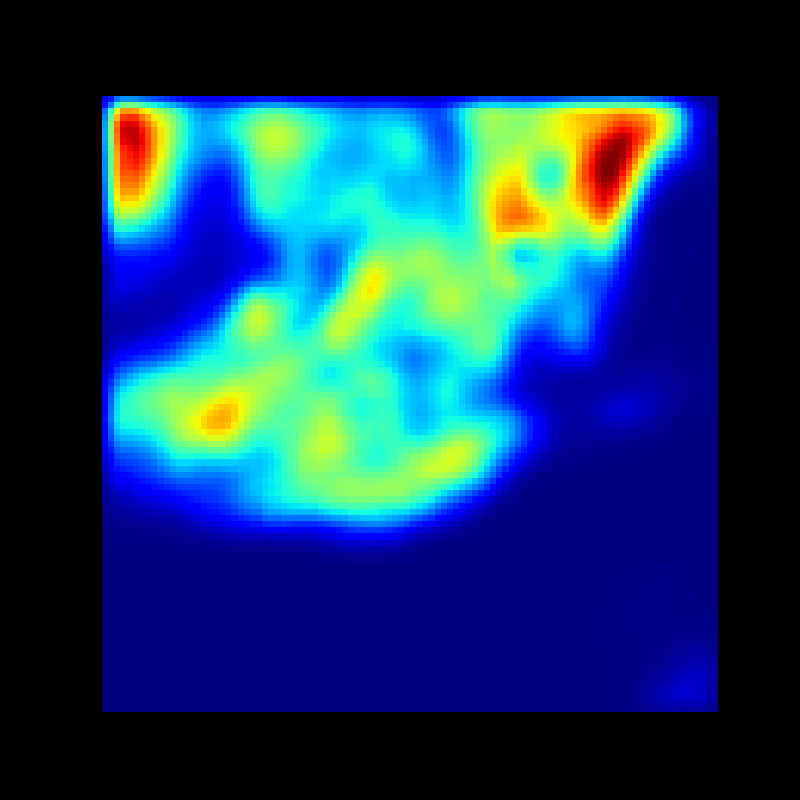} 

	\\
	
	  	\midrule \midrule 
    $t+5$ &
    \includegraphics[height=1.5cm]{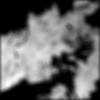} &
    \includegraphics[height=1.5cm]{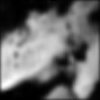} &
    \includegraphics[height=1.5cm]{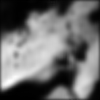} &
    \includegraphics[height=1.5cm,trim={2.5cm 2.5cm 2.5cm 2.5cm},clip]{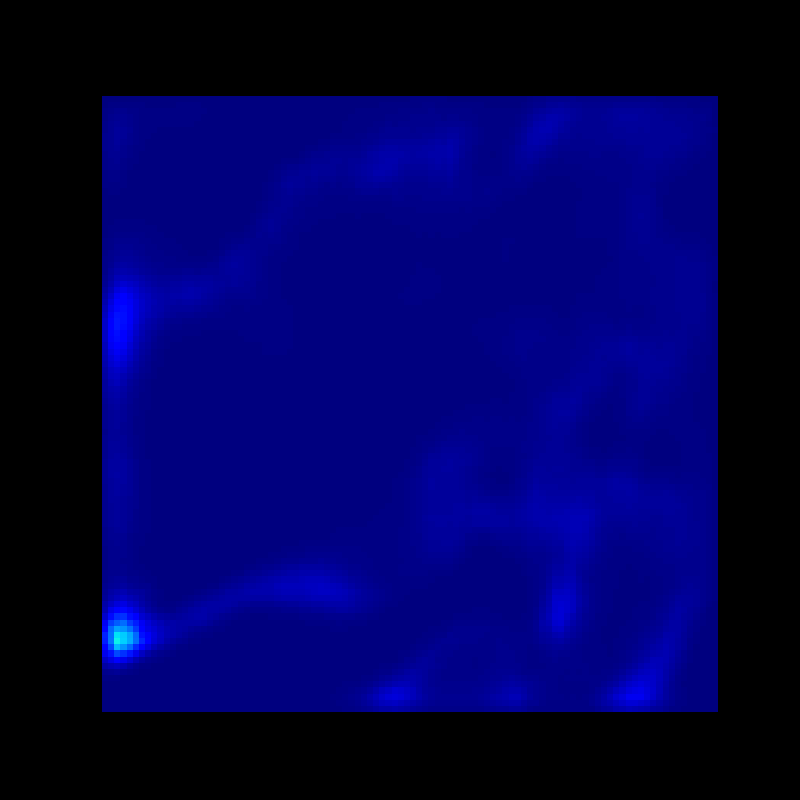} &
    \includegraphics[height=1.5cm]{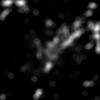} &
    \includegraphics[height=1.5cm]{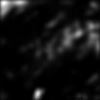} &
    \includegraphics[height=1.5cm]{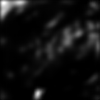} &
    \includegraphics[height=1.5cm,trim={2.5cm 2.5cm 2.5cm 2.5cm},clip]{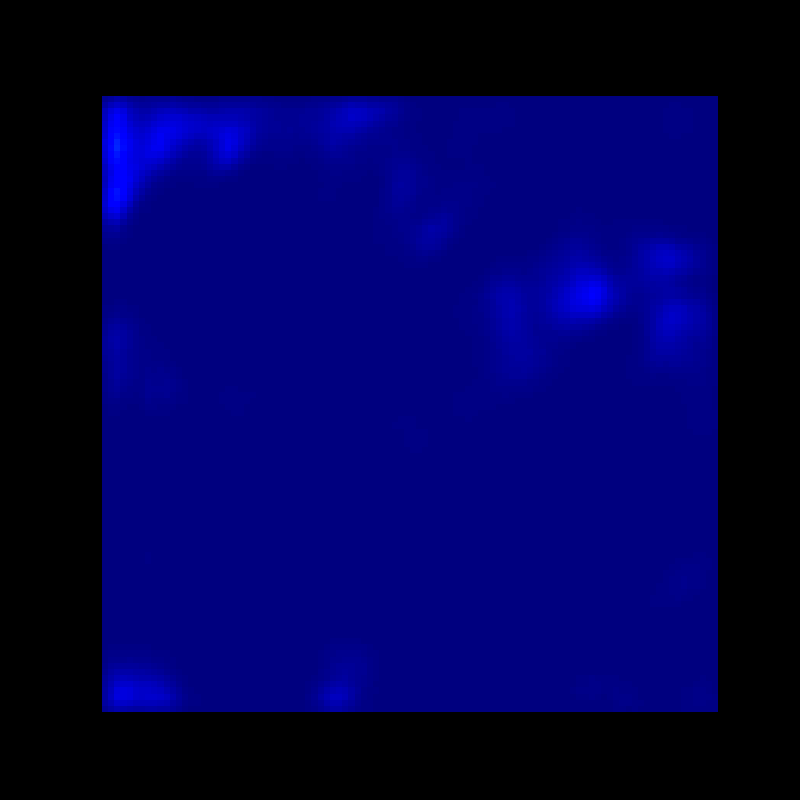} 
    \\
    \midrule
    $t+10$ &
   \includegraphics[height=1.5cm]{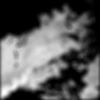} &
    \includegraphics[height=1.5cm]{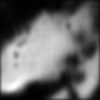} &
    \includegraphics[height=1.5cm]{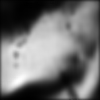} &
    \includegraphics[height=1.5cm,trim={2.5cm 2.5cm 2.5cm 2.5cm},clip]{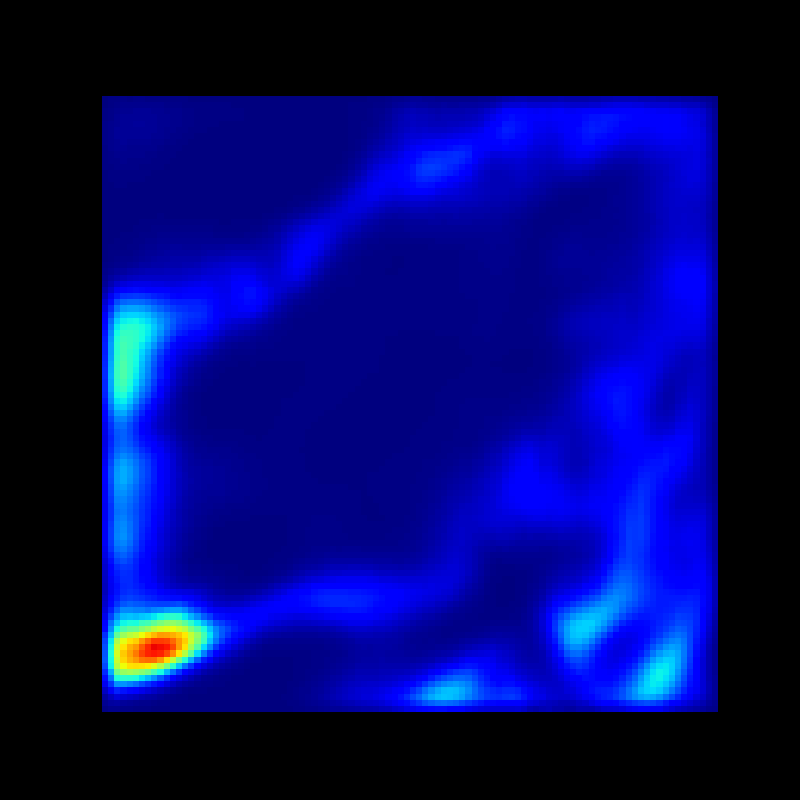} &
    \includegraphics[height=1.5cm]{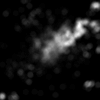} &
    \includegraphics[height=1.5cm]{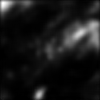} &
    \includegraphics[height=1.5cm]{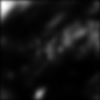} &
    \includegraphics[height=1.5cm,trim={2.5cm 2.5cm 2.5cm 2.5cm},clip]{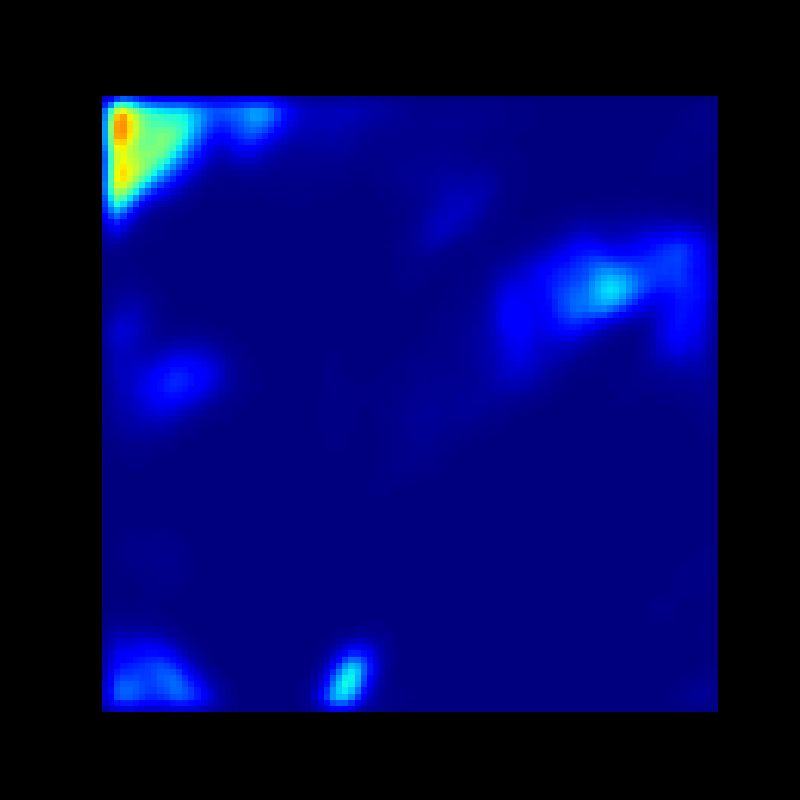} 
    
    \\
    \midrule
    $t+15$ &
    \includegraphics[height=1.5cm]{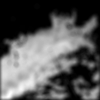} &
    \includegraphics[height=1.5cm]{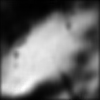} &
    \includegraphics[height=1.5cm]{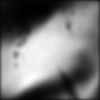} &
    \includegraphics[height=1.5cm,trim={2.5cm 2.5cm 2.5cm 2.5cm},clip]{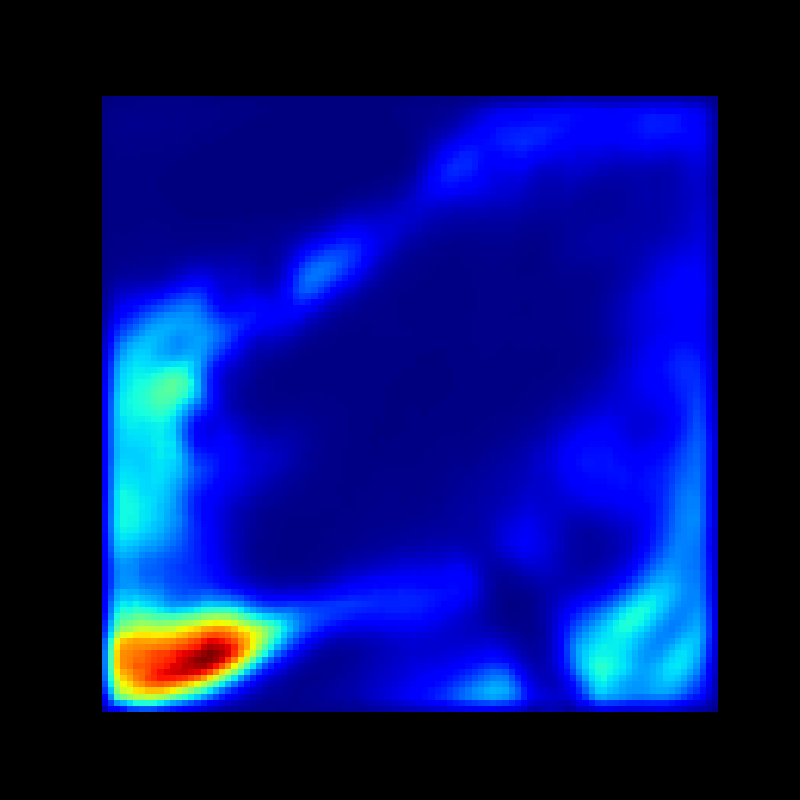} &
    \includegraphics[height=1.5cm]{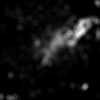} &
    \includegraphics[height=1.5cm]{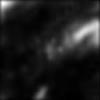} &
    \includegraphics[height=1.5cm]{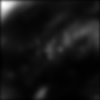} &
    \includegraphics[height=1.5cm,trim={2.5cm 2.5cm 2.5cm 2.5cm},clip]{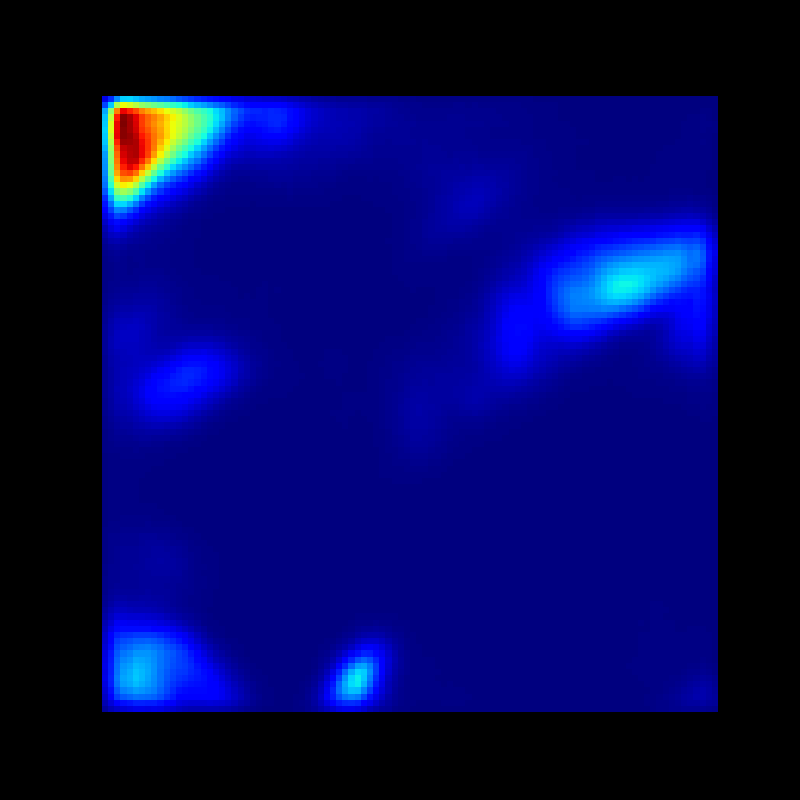} 

	\\
	
	\midrule \midrule
    $t+5$ &
    \includegraphics[height=1.5cm]{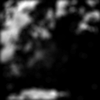} &
    \includegraphics[height=1.5cm]{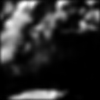} &
    \includegraphics[height=1.5cm]{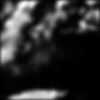} &
    \includegraphics[height=1.5cm,trim={2.5cm 2.5cm 2.5cm 2.5cm},clip]{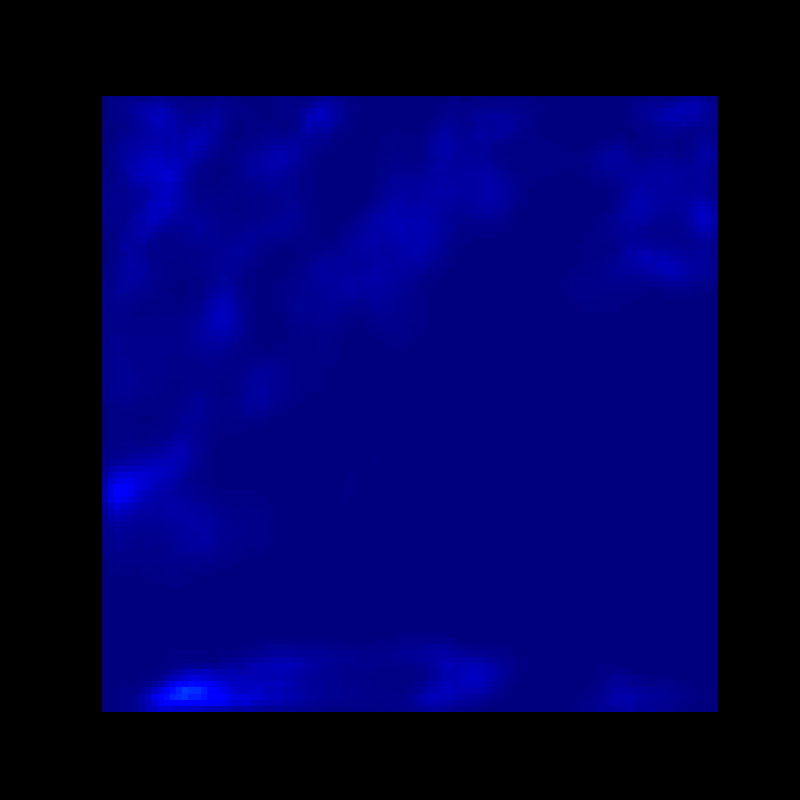} &
    \includegraphics[height=1.5cm]{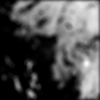} &
    \includegraphics[height=1.5cm]{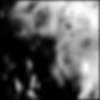} &
    \includegraphics[height=1.5cm]{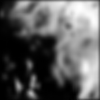} &
    \includegraphics[height=1.5cm,trim={2.5cm 2.5cm 2.5cm 2.5cm},clip]{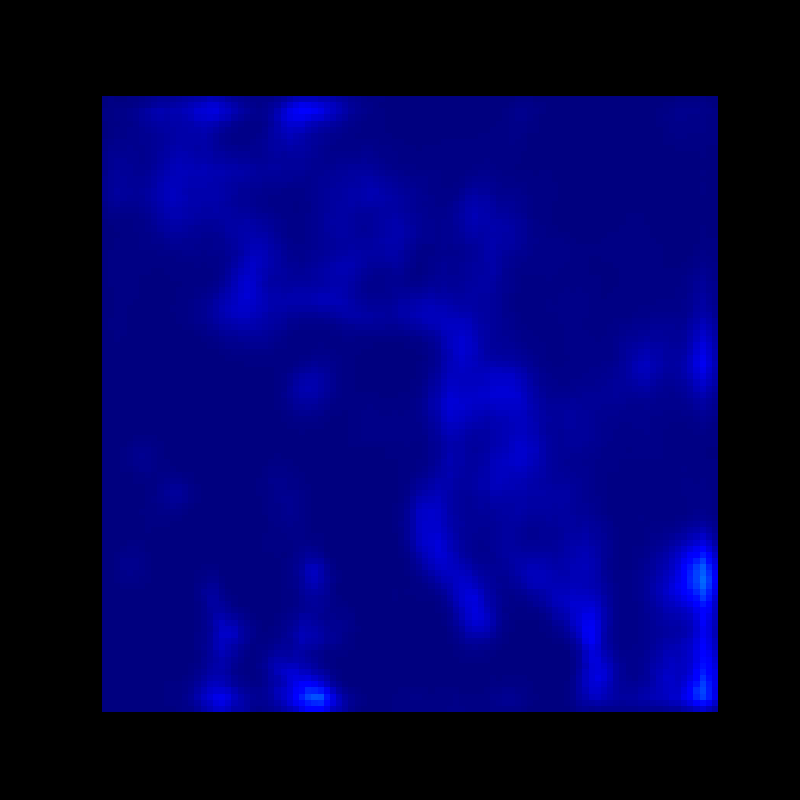} 
    \\
    \midrule
    $t+10$ &
   \includegraphics[height=1.5cm]{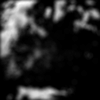} &
    \includegraphics[height=1.5cm]{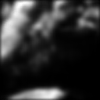} &
    \includegraphics[height=1.5cm]{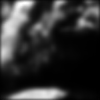} &
    \includegraphics[height=1.5cm,trim={2.5cm 2.5cm 2.5cm 2.5cm},clip]{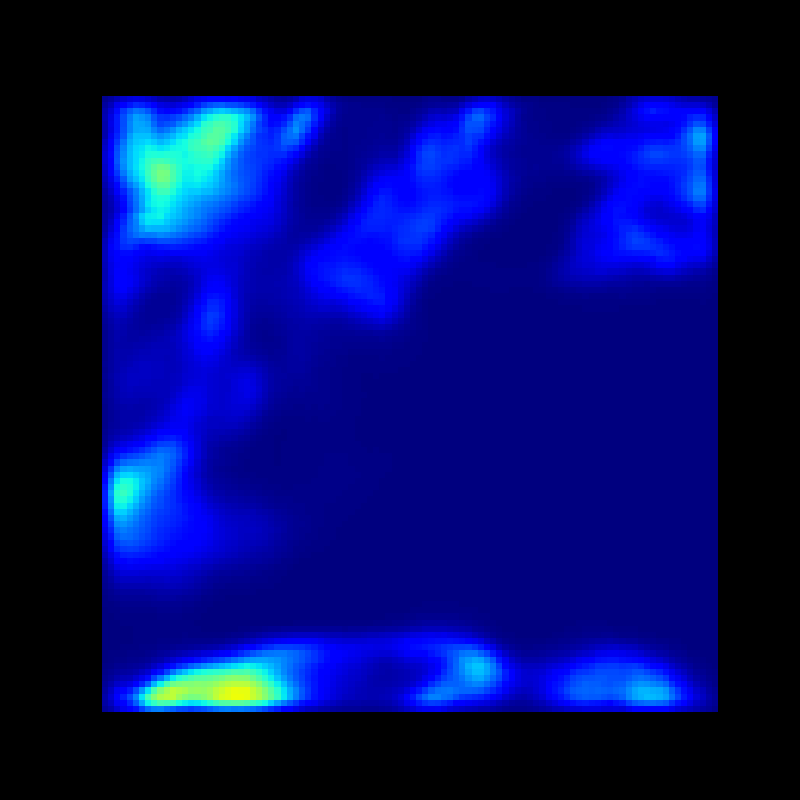} &
    \includegraphics[height=1.5cm]{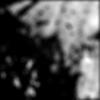} &
    \includegraphics[height=1.5cm]{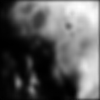} &
    \includegraphics[height=1.5cm]{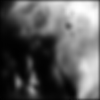} &
    \includegraphics[height=1.5cm,trim={2.5cm 2.5cm 2.5cm 2.5cm},clip]{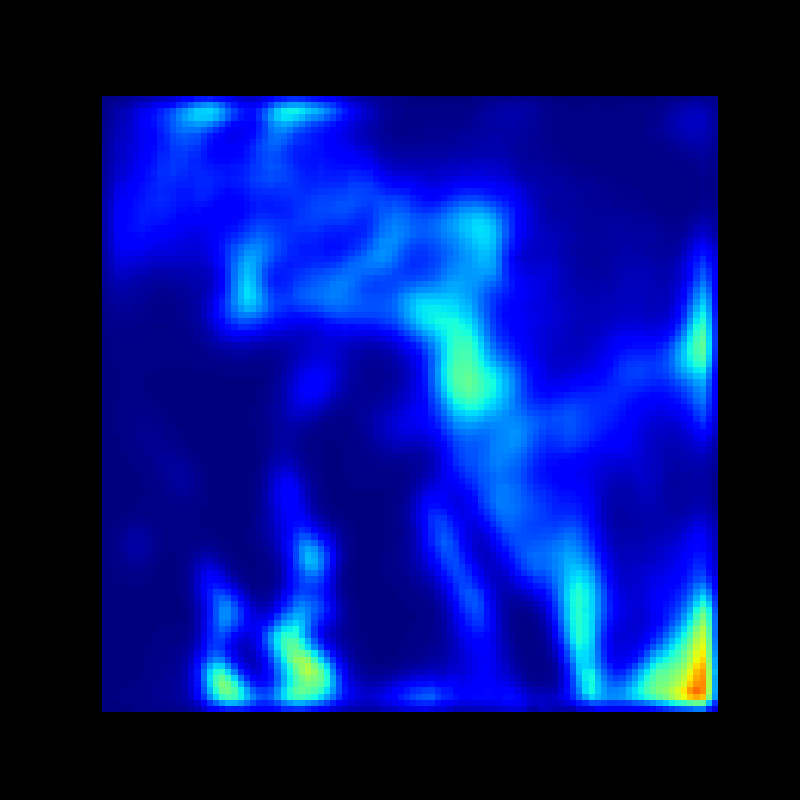} 
    
    \\
    \midrule
    $t+15$ &
    \includegraphics[height=1.5cm]{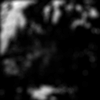} &
    \includegraphics[height=1.5cm]{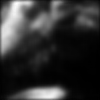} &
    \includegraphics[height=1.5cm]{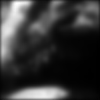} &
    \includegraphics[height=1.5cm,trim={2.5cm 2.5cm 2.5cm 2.5cm},clip]{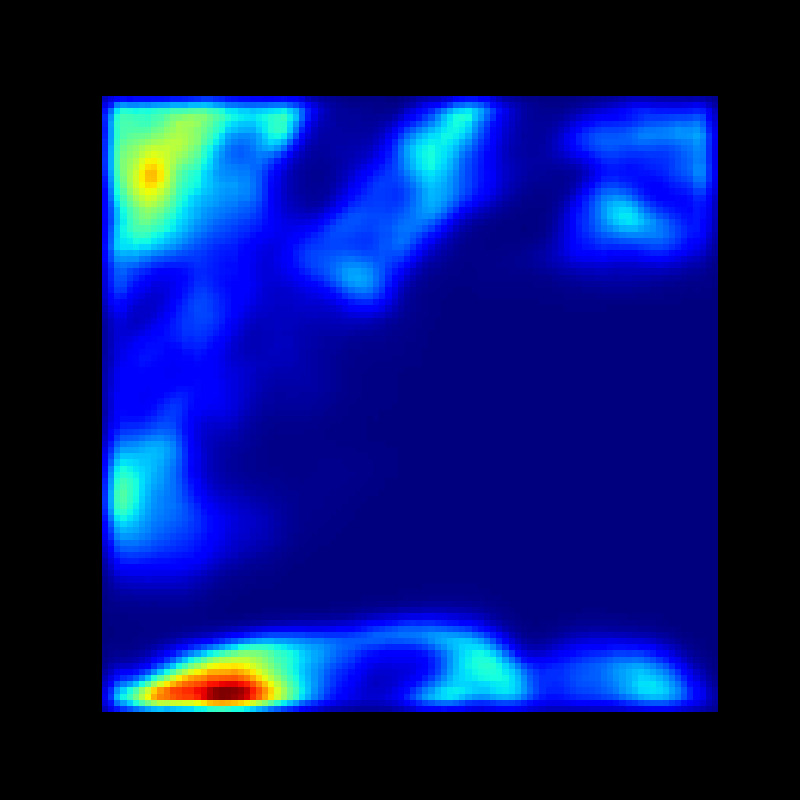} &
    \includegraphics[height=1.5cm]{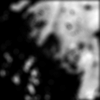} &
    \includegraphics[height=1.5cm]{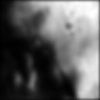} &
    \includegraphics[height=1.5cm]{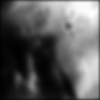} &
    \includegraphics[height=1.5cm,trim={2.5cm 2.5cm 2.5cm 2.5cm},clip]{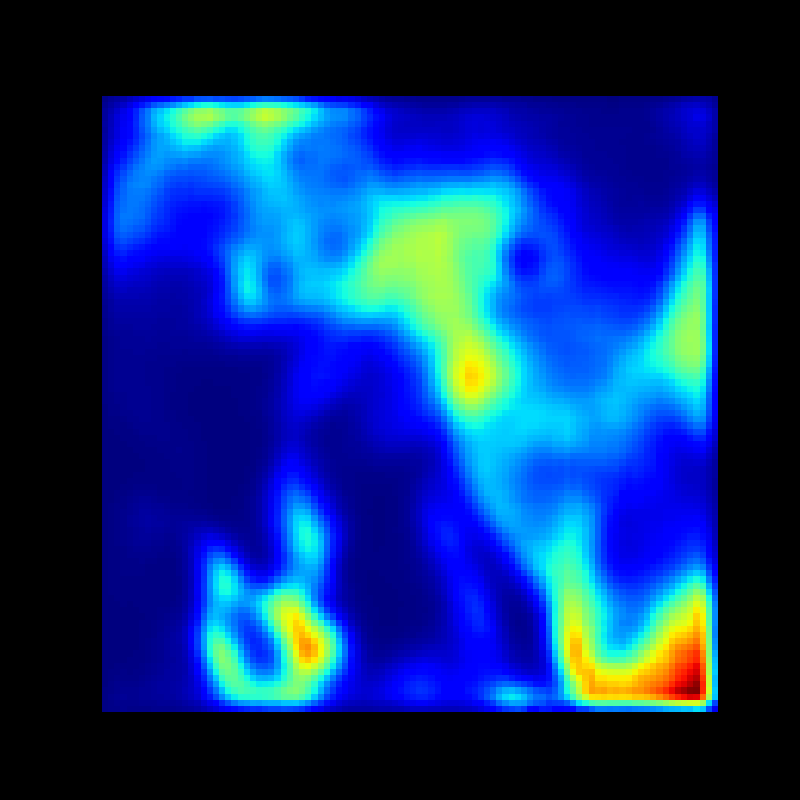} 

	\\
    
    \bottomrule \bottomrule 
    
    \end{tabular}
  \caption{Statistics of samples generated by our LSTM-BMS model on the HKO dataset.}
  \label{fig:hko_frames}
\end{figure*}

\end{appendix}

\end{document}